\documentclass[final,3p,times]{elsarticle}

\usepackage{amsmath}
\usepackage{amssymb}
\usepackage{amsfonts}
\usepackage{graphicx}
\usepackage{booktabs}
\usepackage{multirow}
\usepackage{array}
\usepackage{xcolor}
\usepackage{url}

\journal{Neurocomputing}

\newcommand{\best}[1]{\textbf{#1}}
\newcommand{\second}[1]{\underline{#1}}
\newcommand{\conv}{\circledast}
\newcommand{\had}{\odot}

\begin{document}

\begin{frontmatter}

\title{SFMformer: A Spatial--Frequency Modulation Transformer for
Lightweight Image Super-Resolution}

\author[tku]{Chih-Hsiang Yang}
\ead{FILL-IN-BEFORE-UPLOAD}

\author[tku]{Chia-Min Lin}
\ead{FILL-IN-BEFORE-UPLOAD}

\author[tku]{Ching-Yu Tsai}
\ead{FILL-IN-BEFORE-UPLOAD}

\author[tku]{Yung-Che Wang}
\ead{FILL-IN-BEFORE-UPLOAD}

\author[tku]{Jen-Shiun Chiang\corref{cor1}}
\ead{jsken.chiang@gmail.com}
\cortext[cor1]{Corresponding author.}

\address[tku]{Department of Electrical and Computer Engineering, Tamkang
University, No.\ 151, Yingzhuan Rd., Tamsui District, New Taipei City 251301,
Taiwan}

\begin{abstract}
Sparse attention mechanisms, which score all token pairs but propagate only the
strongest, now underpin the most efficient Transformers for lightweight image
super-resolution. This paper observes that sparsification changes what it means
to improve such a network. A dense attention layer has one place where
representation quality matters: the aggregation of attended features. A sparse
layer has two, because the top-$k$ operator first decides \emph{which} tokens
survive and only then decides what to do with them, and a token discarded at the
selection stage cannot be recovered downstream. Selection quality and
aggregation quality are therefore separable targets, addressed by modules
placed before and after the attention respectively. We test this by pairing a
dual-branch spatial enhancement on the input of a progressive focused attention
with a wavelet-domain modulation on its output, forming SFMformer. Measuring
each module alone and jointly over all fifteen benchmark--scale pairs, we
find their gains are not additive: the joint gain exceeds the sum of the
individual gains on nine pairs, and the sign of the discrepancy is predicted by
how much the weaker module contributes on its own ($r=-0.72$), so the two
compound when they relieve different constraints and overlap when they relieve
the same one. Enabling spectral modulation once per block rather than once per
layer retains the effect at roughly one-sixth of its cost, keeping the model
below one million parameters at every scale. SFMformer ranks first on 28 of 30
PSNR/SSIM entries across five benchmarks and three upscaling factors. We report
the cases where the pairing does not help, and deploy the model on a Raspberry~Pi~5 to
confirm the design is practical under tight resource budgets.
\end{abstract}


\end{frontmatter}

\section{Introduction}
\label{sec:intro}

Single image super-resolution (SR) reconstructs a high-resolution (HR) image
from a low-resolution (LR) observation. It is ill-posed---downsampling is
many-to-one, so one LR input is consistent with infinitely many HR images---and
its defining difficulty is recovering the high-frequency content that
degradation destroys.

Deep learning has driven steady progress, from SRCNN \cite{dong2016srcnn} through
residual and attention-based CNNs \cite{lim2017edsr,zhang2018rdn,zhang2018rcan}
to window-based Transformers \cite{liang2021swinir,chen2023hat}, but the
strongest models need millions of parameters and are impractical where no
accelerator is available. Work on \emph{lightweight} SR therefore targets a
budget below one million parameters, the operating point at which the standard
efficiency benchmarks are constructed \cite{hui2019imdn,luo2020latticenet,
zhang2022elan,zhang2024hitsr,li2022ntireesr,li2023ntireesr,kong2022rlfn}, and it
is the budget we adopt: SFMformer uses 0.97M, 0.98M and 0.99M parameters at
$\times2$, $\times3$ and $\times4$.

The constraint that motivates this is rarely memory alone. Three deployment
patterns recur. In \emph{bandwidth-limited acquisition}---remote cameras, drone
and satellite links, industrial sensors---LR frames are transmitted because the
channel cannot carry more, and reconstruction happens on whatever processor sits
at the receiving end. In \emph{on-device inspection}, an operator examines
imagery on embedded hardware with no accelerator and no network, so latency is
bounded by what a person will wait for. In \emph{archival review}, throughput
matters more than latency but the model must run unattended on commodity
hardware. What these share is not a parameter count but the absence of a GPU,
which is why the sub-1M regime is worth treating as a design target rather than
a leaderboard category; Section~\ref{sec:exp_scenarios} returns to these
patterns with measured timings.

\subsection{Sparse attention has two improvable stages}
\label{sec:intro_contrib}

Among lightweight Transformers, the most efficient now make attention
\emph{sparse}: they score token pairs but propagate only the strongest. PFT
\cite{long2025pft}, our backbone, retains for each query the $K^{l}$ largest
entries of the attention map, with $K^{l}=\alpha K^{l-1}$ so the retained set
contracts with depth, and inherits the map from layer to layer. Sparsification
is usually presented as an efficiency measure. We argue it also changes what it
means to improve such a network.

In a dense attention layer every token contributes to every output and the
projection determines only \emph{how strongly}. An improvement applied before
the attention and one applied after it therefore act on the same quantity---the
weighted combination---and there is little reason to expect them to do
separable work.

Sparsification breaks this symmetry, because the layer now \emph{selects} a
subset of tokens and only then \emph{aggregates} them. These fail differently.
Selection fails when the projected features are not separable enough for the
correct tokens to rank highest, and because $A^{l}$ is inherited by layer
$l+1$, a token dropped early is unavailable to every later layer---the error is
unrecoverable downstream. Aggregation fails when the retained tokens are
combined into a representation whose high-frequency content is too weak for fine
detail, which is a property of the output, not the selection. The two natural
places to intervene are thus not interchangeable: acting on the \emph{input} of
the projection changes the membership of the attended set, acting on its
\emph{output} changes what the aggregated representation contains. We test
whether these are complementary rather than redundant.

We instantiate the two positions with existing components---dual-branch spatial
enhancement \cite{zhang2024hitsr} before the query--key--value projection, and
wavelet-domain modulation \cite{li2025dmnet} after the attention---precisely
because the claim concerns their placement rather than their internal design.
Our contributions are:

\begin{itemize}
\item We identify the selection stage of sparse attention as an intervention
      point distinct from aggregation, and argue it exists only under
      sparsification. Existing work places feature enhancement before attention
      without distinguishing sparse from dense backbones.
\item We characterise how the two interventions interact, including where they
      fail to. Across fifteen benchmark--scale pairs the joint gain exceeds the
      sum of the parts on nine---on Urban100 at $\times2$, $+0.11$ and
      $+0.01$~dB separately but $+0.17$~dB together---and falls below it on six,
      all pairs where both are individually strong. The interaction correlates
      with the weaker intervention's solo gain at $r=-0.72$.
      Section~\ref{sec:exp_interaction} states the criterion this yields and
      the alternative explanation it does not exclude.
\item We show the effect survives aggressive cost reduction: spectral modulation
      applied once per block rather than once per layer retains the benefit at
      roughly one-sixth the cost, keeping the model under 1M parameters.
\item SFMformer ranks first on 28 of 30 PSNR/SSIM entries, with gains
      concentrated on Urban100 and Manga109 as the account predicts. We report
      the configurations where the advantage does not hold, and deploy the model
      on a Raspberry~Pi~5 with an interactive interface.
\end{itemize}

\section{Related work}
\label{sec:related}

\subsection{From CNNs to sparse Transformers}
\label{sec:rel_cnn}

SRCNN \cite{dong2016srcnn} established the deep-learning paradigm for SR with a
three-layer network trained end to end; FSRCNN \cite{dong2016fsrcnn} and ESPCN
\cite{shi2016espcn} then moved upsampling to the end of the network via
sub-pixel convolution, a convention still universal today, and later CNNs
deepened the trunk through residual learning \cite{lim2017edsr}, dense
connectivity \cite{zhang2018rdn} and channel attention \cite{zhang2018rcan}.
The locality of convolution nonetheless caps long-range modelling.

SwinIR \cite{liang2021swinir} addressed this by adapting the shifted-window
self-attention of Swin Transformer \cite{liu2021swin} to restoration, and its
three-stage layout---shallow extraction, deep extraction, reconstruction---has
become the standard skeleton, adopted by HAT \cite{chen2023hat}, SRFormer
\cite{zhou2023srformer} and others. Subsequent work has largely pursued
efficiency: ELAN \cite{zhang2022elan} and OmniSR \cite{wang2023omnisr} reduce
the cost of long-range attention, Restormer \cite{zamir2022restormer} moves
attention to the channel axis, and MambaIR \cite{guo2024mambair} replaces it
with a state-space model.

The most relevant strand makes attention \emph{sparse}. ATD \cite{zhang2024atd}
replaces exhaustive pairwise similarity with a learnable token dictionary, and
PFT \cite{long2025pft}---our backbone---introduces Progressive Focused Attention
(PFA), which inherits the attention map of the previous layer, multiplies it
element-wise with the current similarity, and keeps only the top $K^{l}$
entries, with $K^{l}=\alpha K^{l-1}$ so that the retained set contracts with
depth. Shallow layers thus explore broadly and deep layers focus precisely,
while the sparsified index matrix is reused for aggregation, avoiding further
multiplications. Two properties of PFA matter for what follows: $Q$, $K$ and $V$
come from a purely channel-wise linear projection that cannot see local spatial
structure, and the mechanism has no representation of frequency content.

\subsection{Feature enhancement and frequency-domain modelling}
\label{sec:rel_modules}

Two lines of work supply the components we place around this backbone. The
first strengthens features before attention: HiT-SR \cite{zhang2024hitsr}
observes that a linear QKV projection ignores spatial neighbourhood structure
and prepends a Dual Feature Extraction (DFE) module, whose hourglass
convolutional branch ($1\times1\rightarrow3\times3\rightarrow1\times1$) captures
local texture while a parallel $1\times1$ branch preserves channel semantics,
the two combining by element-wise product. HiT-SR applies it to \emph{dense}
self-correlation attention.

The second models the frequency domain explicitly, on the reasoning that SR is
fundamentally the recovery of high-frequency content. Wavelet-domain methods
\cite{xin2022wdrn,zou2023jwsgn,jiang2023fabnet} separate low- and
high-frequency sub-bands for independent treatment; Fourier-domain methods
exploit the convolution theorem for efficiency \cite{kong2023fftformer,li2023fsr}
or combine spectra with state-space models \cite{xiao2025freqmamba}. Closest to
our use is DMNet \cite{li2025dmnet}, whose Wavelet-domain Modulation
self-Attention (WMA) decomposes features by DWT into LL/LH/HL/HH sub-bands,
computes attention \emph{across} the concatenated sub-bands so that frequency
bands can inform one another, refines the result with a dynamic convolution and
returns to the spatial domain by IDWT. DMNet places WMA alongside a
channel-attention spatial branch, with no sparse selection anywhere in the
model. Others position a spectral prior differently again: SwinFIR
\cite{zhang2022swinfir} attaches a spatial--frequency block to the tail of the
whole trunk, and FreqFormer \cite{dai2024freqformer} folds frequency awareness
into the attention operator itself.

Read together, these methods act on the two stages identified in
Section~\ref{sec:intro_contrib} without connecting them. DFE sits before an
attention that is dense, so it can only reweight an attended set whose
membership is fixed; WMA sits after an attention with no selection stage
upstream to be complementary to; and PFT supplies the sparsification that makes
the two positions distinct, yet leaves both empty. The configuration studied
here---selection-side enhancement and aggregation-side modulation around a
progressively sparsified attention---has therefore not been examined.

\section{Proposed method}
\label{sec:method}

\subsection{Overall architecture}
\label{sec:method_overall}

\begin{figure*}[t]
\centering
\includegraphics[width=\textwidth]{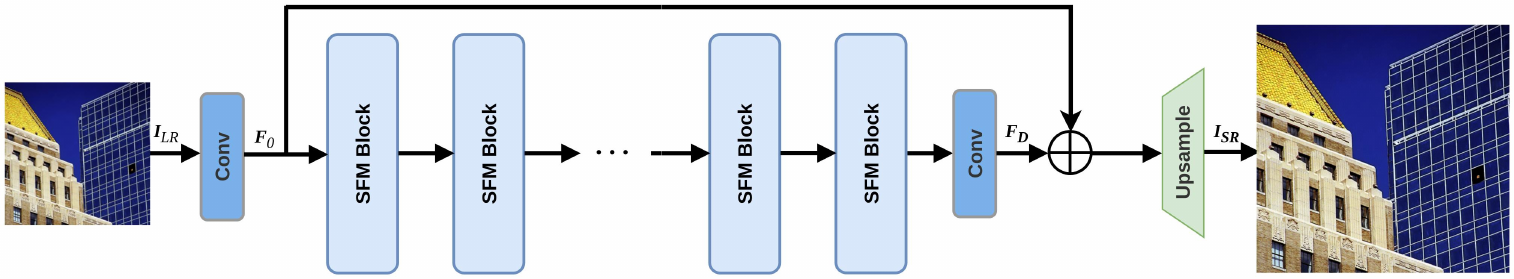}
\caption{Overall architecture of SFMformer. A $3\times3$ convolution extracts
shallow features, $M$ cascaded SFM Blocks produce deep features, and a long
residual connection followed by pixel-shuffle upsampling reconstructs the HR
output.}
\label{fig:arch}
\end{figure*}

SFMformer acts on the two stages identified in
Section~\ref{sec:intro_contrib} on a PFT \cite{long2025pft} backbone. The
selection-side position is filled by a DFE module \cite{zhang2024hitsr} placed
before every query--key--value projection, so that the features on which the
top-$k$ operator ranks tokens carry explicit local spatial structure. The
aggregation-side position is filled by a WMA module \cite{li2025dmnet} placed
after the attention, where it redistributes energy across frequency sub-bands of
the already-aggregated representation. The two are deliberately asymmetric in
cost: DFE runs in every layer because selection happens in every layer, whereas
WMA runs only in the final SFM Layer of each SFM Block, on the reasoning that
the aggregated representation need be spectrally corrected once per block rather
than once per layer. The architecture is shown in Fig.~\ref{fig:arch}.

Given an LR image $I_{LR}$, an initial $3\times3$ convolution extracts shallow
features $F_{0}\in\mathbb{R}^{C\times H\times W}$ as in Eq.~\eqref{eq:shallow};
$M$ cascaded SFM Blocks then perform deep feature extraction to yield $F_{D}$
as in Eq.~\eqref{eq:deep}; finally a residual connection adds the shallow and
deep features and an upsampling module restores the image to $I_{SR}$ as in
Eq.~\eqref{eq:upsample}:

\begin{align}
F_{0} &= \mathrm{Conv}_{3\times3}(I_{LR}), \label{eq:shallow}\\
F_{D} &= \mathcal{H}_{SFMB}(F_{0}), \label{eq:deep}\\
I_{SR} &= \mathrm{Upsample}\bigl(\mathrm{Conv}_{3\times3}(F_{D}) + F_{0}\bigr).
\label{eq:upsample}
\end{align}

\subsection{Spatial--Frequency Modulation Block (SFMB)}
\label{sec:method_sfmb}

\begin{figure}[t]
\centering
\includegraphics[width=\columnwidth]{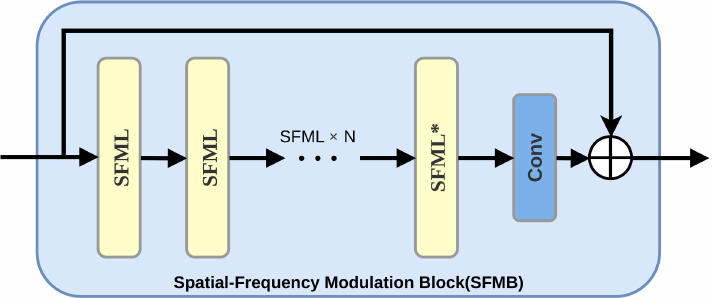}
\caption{Structure of the Spatial--Frequency Modulation Block (SFMB). Only the
final layer (SFML$^{*}$) enables the WMA module.}
\label{fig:sfmb}
\end{figure}

The SFM Block is the core deep-feature unit of SFMformer. Each SFMB consists of
$N$ cascaded Spatial--Frequency Modulation Layers (SFML) followed by a
$3\times3$ convolution and a short residual connection. The last layer, denoted
SFML$^{*}$, additionally enables the WMA module for global frequency-domain
modulation; the remaining layers do not, and perform spatial-domain feature
modelling only. The block is illustrated in Fig.~\ref{fig:sfmb}.

The $3\times3$ convolution follows the standard design of SwinIR
\cite{liang2021swinir}: it supplies the local spatial correlation that the
attention mechanism lacks and performs a final aggregation of the features
extracted by the SFMLs, while the residual connection stabilises training of
the deep network. The computation is given by Eq.~\eqref{eq:sfmb}:

\begin{equation}
\mathbf{F}_{b}^{out} = \mathrm{Conv}_{3\times3}\Bigl(
\mathrm{SFML}_{N}\bigl(\dots\mathrm{SFML}_{1}(\mathbf{F}_{b}^{in})\bigr)\Bigr)
+ \mathbf{F}_{b}^{in},
\label{eq:sfmb}
\end{equation}

where $\mathbf{F}_{b}^{in}$ and $\mathbf{F}_{b}^{out}$ are the input and output
features of the $b$-th SFMB.

\begin{figure*}[t]
\centering
\includegraphics[width=\textwidth]{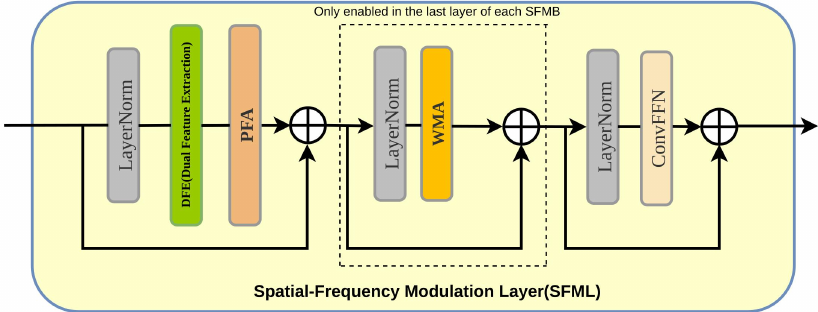}
\caption{Internal structure of the Spatial--Frequency Modulation Layer (SFML).
The dashed region (LayerNorm + WMA) is enabled only in the last layer of each
SFMB.}
\label{fig:sfml}
\end{figure*}

The internal structure of an SFML is shown in Fig.~\ref{fig:sfml} and follows
the standard three-stage Transformer block layout. In the first stage the input
is normalised, enhanced by the DFE module \cite{zhang2024hitsr}, passed to PFA
\cite{long2025pft} for progressive attention, and added back through a residual
connection. The second stage applies the WMA module \cite{li2025dmnet} for
global frequency-domain modulation, and is active only in the last SFML of each
SFMB. The third stage is a convolutional feed-forward network performing
non-linear feature transformation. Each stage carries its own residual
connection, ensuring that features are refined layer by layer without losing
the original information:

\begin{align}
X' &= X + \mathrm{PFA}\bigl(\mathrm{DFE}(\mathrm{LN}(X))\bigr),
\label{eq:sfml1}\\
X'' &= \begin{cases}
X' + \mathrm{WMA}(\mathrm{LN}(X')), & \text{if last layer of the block},\\[2pt]
X', & \text{otherwise},
\end{cases}
\label{eq:sfml2}\\
X_{out} &= X'' + \mathrm{ConvFFN}(\mathrm{LN}(X'')).
\label{eq:sfml3}
\end{align}

\subsubsection{Dual Feature Extraction (DFE)}
\label{sec:method_dfe}

As described above, PFA \cite{long2025pft} propagates attention maps across
layers and selects the top-$k$ most important tokens for each query. This
mechanism depends heavily on the spatial discriminability of $Q$ and $K$: if
neighbouring tokens differ only slightly, the selection cannot converge on the
genuinely critical positions. Yet $Q$, $K$ and $V$ are generated by a purely
channel-wise linear projection, so the LayerNorm-ed features entering PFA carry
no explicit spatial structural signature.

To resolve this we embed the dual-branch feature extraction module of HiT-SR
\cite{zhang2024hitsr} in front of the QKV projection. The original authors
present DFE as a general pre-attention
enhancement module, placed before their spatial (S-SC) and channel (C-SC)
self-correlation attention. We instead introduce it into the window-based
\emph{sparse} attention framework of PFT, so that top-$k$ selection is
performed on spatially discriminative $Q$/$K$/$V$ and attention converges more
precisely. A second difference is that HiT-SR splits the DFE output into $Q$
and $V$ only (with $K$ equal to $V$) to suit its self-correlation design,
whereas we retain the standard three-way $Q$, $K$, $V$ projection of PFT in
order to preserve the independent key representation that PFA requires when
forming its attention map.

\begin{figure}[t]
\centering
\includegraphics[width=\columnwidth]{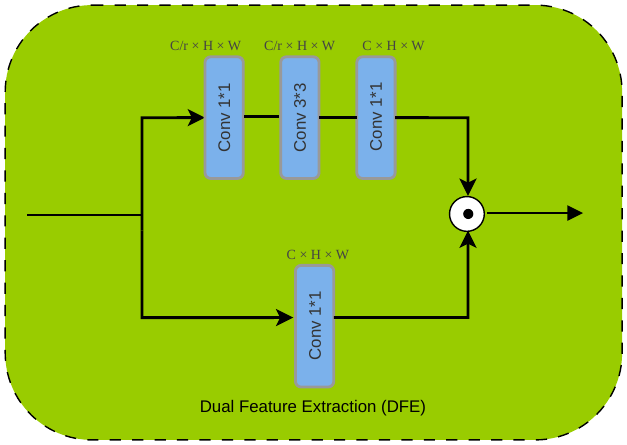}
\caption{The Dual Feature Extraction (DFE) module. The hourglass branch captures
local spatial neighbourhood information; the $1\times1$ branch preserves channel
semantics.}
\label{fig:dfe}
\end{figure}

The module is shown in Fig.~\ref{fig:dfe} and defined by
$\mathrm{DFE}(X) = (W_{g}\conv X)\had
\bigl(W_{3}\conv\sigma(W_{2}\conv\sigma(W_{1}\conv X))\bigr)$, where $\conv$ is
convolution and $\had$ element-wise product. The first factor is the channel
branch, a $1\times1$ convolution equivalent to an independent linear projection
at every spatial position; the second is the spatial branch, whose
$1\times1\rightarrow3\times3\rightarrow1\times1$ bottleneck extracts local
neighbourhood structure. Following HiT-SR the bottleneck ratio is $r=5$ and
$\sigma$ is LeakyReLU with negative slope $0.2$.

\subsubsection{Progressive Focused Attention (PFA)}
\label{sec:method_pfa}

After DFE enhancement the features enter the attention layer. SFMformer adopts
PFA \cite{long2025pft} as its core attention operator, performing efficient
sparse attention within each window through a cross-layer top-$k$
sparsification strategy. As described in Section~\ref{sec:rel_cnn}, PFA
rests on two mechanisms: passing the attention distribution of the previous
layer to the current one so that importance judgements are consistent across
consecutive layers, and using the inherited scores to decide which tokens
should still participate, retaining only the most representative few through
top-$k$ sparsification. This reduces redundant computation and suppresses noise
from unrelated features.

Given the input feature $X^{l}$ of layer $l$, the attention computation of an
SFM Layer is described by Eqs.~\eqref{eq:qkv}--\eqref{eq:agg}:

\begin{align}
[Q^{l},K^{l},V^{l}] &= W^{l}_{qkv}\bigl(\mathrm{DFE}(X^{l})\bigr),
\label{eq:qkv}\\
A^{l} &= \mathrm{Top}_{K^{l}}\!\left(
\mathrm{Softmax}\!\left(\frac{Q^{l}(K^{l})^{\top}}{\sqrt{d}}\right)\had A^{l-1}
\right), \label{eq:pfa}\\
Z^{l} &= A^{l}V^{l}. \label{eq:agg}
\end{align}

\begin{figure*}[t]
\centering
\includegraphics[width=\textwidth]{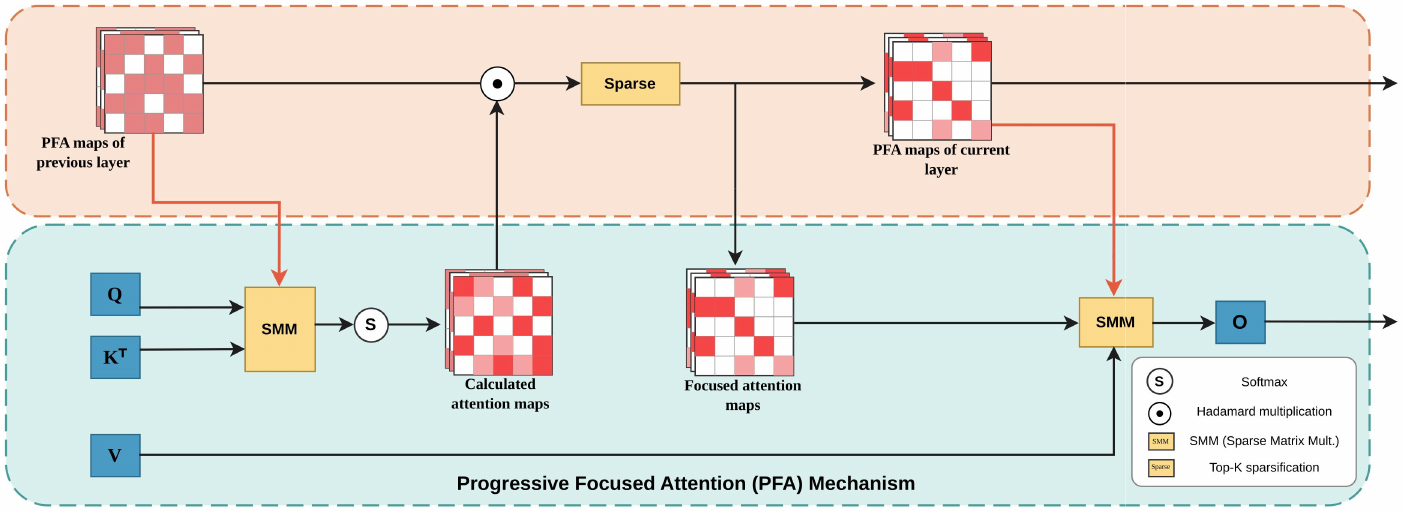}
\caption{The Progressive Focused Attention (PFA) mechanism. The attention map of
the previous layer is inherited, multiplied element-wise with the current
similarity map, and sparsified by a top-$K^{l}$ operation.}
\label{fig:pfa}
\end{figure*}

Equation~\eqref{eq:qkv} is the attention pre-processing stage: the input is
first enhanced by the DFE module of Section~\ref{sec:method_dfe} and then
mapped by the linear projection $W^{l}_{qkv}$ into the three tensors $Q$, $K$
and $V$. Equation~\eqref{eq:pfa} is the PFA operation itself: after the
similarity between the current $Q^{l}$ and $K^{l}$ has been computed, it is
multiplied element-wise with the attention map $A^{l-1}$ inherited from the
previous layer, realising cross-layer attention inheritance, and a top-$K^{l}$
sparsification produces the attention map $A^{l}$ of the current layer. This
design allows subsequent attention layers to identify which features matter and
amplify their weights, while unrelated features are suppressed. Because $A^{l}$
retains only $K^{l}$ non-zero values, the next layer need compute similarity
only at those non-zero positions. The map $A^{l}$ is then passed on as the
input to the PFA operation of the following layer, forming a cross-layer
attention chain, with $K^{l}=\alpha K^{l-1}$ so that the retained token count
decays with depth. The mechanism is illustrated in Fig.~\ref{fig:pfa}.

\subsubsection{Wavelet-domain Modulation self-Attention (WMA)}
\label{sec:method_wma}

\begin{figure*}[t]
\centering
\includegraphics[width=\textwidth]{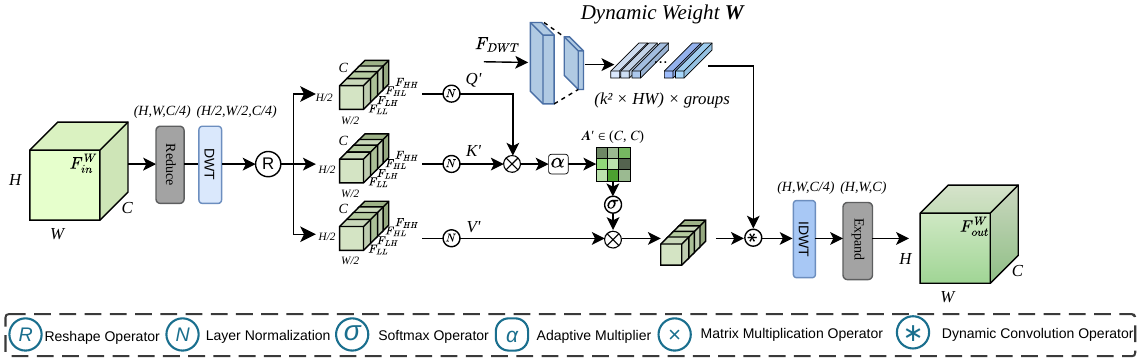}
\caption{The Wavelet-domain Modulation self-Attention (WMA) module. Features
are reduced in channel dimension, decomposed by the DWT into four sub-bands,
modulated by channel-wise self-attention and dynamic convolution, and returned
to the spatial domain by the IDWT.}
\label{fig:wma}
\end{figure*}

To compensate for the limitations of purely spatial modelling we combine
frequency- and spatial-domain features. Concretely we adopt the Wavelet-domain
Modulation self-Attention module (WMA) \cite{li2025dmnet}, which applies global
modulation in the frequency domain to the features produced by PFA
\cite{long2025pft}.
For computational efficiency, WMA is not used in every SFM Layer but only in
the last SFM Layer of each SFMB; its operation is shown in
Fig.~\ref{fig:wma}.

When the PFA output is passed to WMA the feature is first reduced in channel
dimension to lower the computational cost, and is then decomposed by the
discrete wavelet transform into four sub-bands, each of size
$(H/2,\,W/2,\,C/4)$, as in Eq.~\eqref{eq:dwt}:

\begin{equation}
\{F_{LL},F_{LH},F_{HL},F_{HH}\} = \mathrm{DWT}\bigl(\mathrm{Conv}_{1\times1}
(F_{in}^{W})\bigr).
\label{eq:dwt}
\end{equation}

The sub-bands are arranged along the channel dimension and reshaped, after
which the similarity between the resulting vectors is computed as in
Eqs.~\eqref{eq:wma_qkv}--\eqref{eq:wma_att}:

\begin{align}
[Q',K',V'] &= \mathrm{LN}\bigl(\mathcal{R}(\{F_{LL},F_{LH},F_{HL},F_{HH}\})
\bigr), \label{eq:wma_qkv}\\
A' &= \mathrm{Softmax}\bigl(\alpha\, Q'(K')^{\top}\bigr) \in \mathbb{R}^{C
\times C}, \label{eq:wma_att}
\end{align}

where $\mathcal{R}(\cdot)$ denotes the reshape operator and $\alpha$ an
adaptive multiplier. Because the attention matrix is formed across the channel
axis on the concatenated sub-bands, correlations are established \emph{between}
frequency bands, allowing low- and high-frequency information to reinforce one
another. Once the similarity has been computed, a dynamic convolution weight
$W$ injects spatial information into the wavelet-domain feature, strengthening
its representation; the modulated feature is finally returned to the spatial
domain by the inverse wavelet transform, as in Eq.~\eqref{eq:idwt}:

\begin{equation}
F_{out}^{W} = \mathcal{E}\Bigl(\mathrm{IDWT}\bigl((A'V')\ast W\bigr)\Bigr),
\label{eq:idwt}
\end{equation}

where $\ast$ denotes the dynamic convolution operator and $\mathcal{E}(\cdot)$
the channel-expansion operation that restores the original width.

\subsection{High-quality image reconstruction}
\label{sec:method_recon}

Once deep feature extraction is complete the features must be reconstructed
into a high-quality image. As described in Section~\ref{sec:rel_cnn},
sub-pixel convolution \cite{shi2016espcn} rearranges the pixels of each channel
of the input feature map into the spatial dimensions of the output. We use
\emph{pixel-shuffle direct}, the variant provided for lightweight models in the
official SwinIR implementation \cite{liang2021swinir}. Relative to the standard
version it omits the preceding feature pre-processing convolution and the
multi-stage upsampling procedure, integrating them into a single convolution,
as in Eq.~\eqref{eq:recon}:

\begin{equation}
I_{SR} = \mathrm{PixelShuffle}_{r}\bigl(W_{up}\conv F_{deep}\bigr),
\label{eq:recon}
\end{equation}

where $F_{deep}\in\mathbb{R}^{C\times H\times W}$ is the output of the deep
feature extraction stage and $W_{up}\in\mathbb{R}^{(r^{2}\cdot3)\times C\times
3\times3}$ is a single upsampling convolution mapping the channel count from
$C$ directly to $r^{2}\cdot3$, with $r$ the upscaling factor.
$\mathrm{PixelShuffle}_{r}(\cdot)$ rearranges the $r^{2}$-fold channel
dimension into the spatial dimensions, producing the final RGB image
$I_{SR}\in\mathbb{R}^{3\times rH\times rW}$. Compared with the standard
version this markedly reduces the parameters and computation of the
reconstruction stage, helping to lower model complexity while maintaining
quality.

\subsection{Loss function}
\label{sec:method_loss}

Training uses a pixel-domain loss ($L_{1}$) together with a frequency-domain
loss ($L_{fft}$), supervising reconstruction quality in the spatial and
frequency domains respectively.

Pixel loss is the most basic supervisory signal in SR, measuring the
pixel-wise difference between the reconstruction $I_{SR}$ and the ground-truth
$I_{HR}$. We use the mean absolute error, also known as $L_{1}$ loss, given by
Eq.~\eqref{eq:l1}:

\begin{equation}
L_{1} = \frac{1}{N}\sum_{i=1}^{N}\bigl|I_{SR}(i) - I_{HR}(i)\bigr|.
\label{eq:l1}
\end{equation}

Compared with $L_{2}$ loss (mean squared error), $L_{1}$ is less sensitive to
outliers and has been shown experimentally to produce sharper reconstructions
\cite{lim2017edsr}, which is why it is the standard choice in the SR
literature.

Because our model combines spatial and frequency features, pixel-domain
supervision alone is insufficient to constrain frequency-domain reconstruction
quality. We therefore adopt the Fourier loss used by DMNet
\cite{li2025dmnet}: the reconstruction $I_{SR}$ and the ground truth $I_{HR}$
are transformed to the Fourier domain by a two-dimensional fast Fourier
transform and their $L_{1}$ distance is computed, as in Eq.~\eqref{eq:fft}:

\begin{equation}
L_{fft} = \bigl\|\,\mathcal{F}(I_{SR}) - \mathcal{F}(I_{HR})\,\bigr\|_{1},
\label{eq:fft}
\end{equation}

where $\mathcal{F}(\cdot)$ denotes the 2D FFT. As noted in
\cite{li2025dmnet}, a Fourier loss is preferred over a wavelet-domain loss
because the gradient descent directions of different wavelet sub-bands can be
markedly inconsistent, and the resulting conflicts make convergence difficult.
Supervising in the Fourier domain avoids these inter-sub-band gradient
conflicts. The wavelet domain is thus responsible for reconstructing
high-frequency texture, while the Fourier domain regulates the global frequency
distribution, jointly optimising overall structure and fine detail.

Combining the two terms, training uses the weighted sum of
Eq.~\eqref{eq:total}:

\begin{equation}
L_{total} = L_{1} + \lambda\cdot L_{fft}.
\label{eq:total}
\end{equation}

Following the default setting of DMNet \cite{li2025dmnet} we set
$\lambda=0.1$. The model therefore receives supervisory signals from both the
spatial and the frequency domain during training, balancing structural
correctness (pixel supervision) against detail sharpness (frequency
supervision).

\section{Experiments}
\label{sec:experiments}

\subsection{Datasets and evaluation protocol}
\label{sec:exp_data}

Following the standard protocol for Transformer-based SR, we train on DF2K---the
union of the 800 DIV2K \cite{agustsson2017div2k} training images and the 2650
Flickr2K \cite{timofte2017ntire} images, 3450 in total---and evaluate on five
benchmarks: Set5 \cite{bevilacqua2012set5}, Set14 \cite{zeyde2010set14}, BSD100
\cite{martin2001bsd}, Urban100 \cite{huang2015urban100} and Manga109
\cite{matsui2017manga109}. LR inputs are produced by bicubic downsampling. The
last two benchmarks are the most informative for this work: Urban100 is
dominated by man-made structures with regular repeating texture and straight
edges, which stress long-range dependency modelling and the exploitation of
self-similarity, while Manga109 consists of comic artwork whose sharp,
high-contrast line work concentrates energy in the high-frequency sub-bands.
They therefore probe the spatial and the spectral half of our design
respectively.

We report PSNR and SSIM \cite{wang2004ssim}, both computed on the Y (luminance)
channel of the YCbCr colour space with the image border cropped by a number of
pixels equal to the upscaling factor, as is conventional. PSNR measures
pixel-wise fidelity and SSIM the preservation of local luminance, contrast and
structure; we use the original formulation and constants of Wang et al.\
\cite{wang2004ssim} without modification.

\subsection{Implementation details}
\label{sec:exp_impl}

All experiments are implemented with PyTorch 2.8.0 and the BasicSR framework.
Training and testing use Windows 11 with an NVIDIA RTX 5090 GPU and CUDA 12.8.
HR patches are randomly cropped from the DF2K training set, with a crop size
that depends on the upscaling factor: $128\times128$ for $\times2$,
$192\times192$ for $\times3$ and $256\times256$ for $\times4$, so that the
corresponding LR input patch is uniformly $64\times64$. Data augmentation
comprises random horizontal flipping and random rotation by $90^{\circ}$,
$180^{\circ}$ and $270^{\circ}$.

We use the Adam optimiser with an initial learning rate of $2\times10^{-4}$,
$\beta_{1}=0.9$ and $\beta_{2}=0.99$. The learning rate follows a multi-step
decay schedule, halving at iterations 250K, 400K, 450K and 475K, with 500K
total iterations. The batch size is 8. The loss is configured as described in
Section~\ref{sec:method_loss}, with the Fourier loss weight $\lambda=0.1$. An
exponential moving average of the weights (decay 0.999) is maintained during
training. The $\times3$ and $\times4$ models are fine-tuned from the trained
$\times2$ weights for 250K iterations each.

\subsection{Comparison with state-of-the-art methods}
\label{sec:exp_sota}

Table~\ref{tab:sota} reports the main quantitative comparison against
representative lightweight Transformers
\cite{liang2021swinir,zhang2022elan,choi2023swinirng,wang2023omnisr,li2025dmnet,
zhou2023srformer,zhang2024hitsr,tian2024ipg,zhang2024atd,hu2025lkmn,
liu2026promptsr,zhang2022swinfir,dai2024freqformer}, with bicubic as a lower
reference excluded from the ranking. Methods marked $\dagger$ were retrained
under our configuration; other figures are quoted from the original papers, and
FLOPs are computed at an output resolution of $1280\times720$. The batch size
here is 8; PFT-light was originally trained at 32, so its retrained figures
differ slightly from the published ones, and the effect of raising the batch
size is examined separately in Section~\ref{sec:exp_bs}.

SFMformer obtains the best result on 28 of the 30 PSNR/SSIM entries and the
second best on the remaining two, both Set14 at $\times2$, where IPG-Tiny
\cite{tian2024ipg} leads; SFMformer leads IPG-Tiny on the same dataset at
$\times3$ and $\times4$, so the gap reflects that benchmark's 14-image sample
rather than a systematic weakness. Against the retrained baseline the
comparison is strictly controlled---identical patch size, optimiser, schedule
and iteration count, differing only in the two proposed modules---and SFMformer
improves on it at every dataset and scale, by $+0.17$/$+0.23$~dB on
Urban100/Manga109 at $\times2$ and $+0.15$/$+0.20$~dB at $\times4$.

\begin{figure*}[t]
\centering
\includegraphics[width=0.8\textwidth]{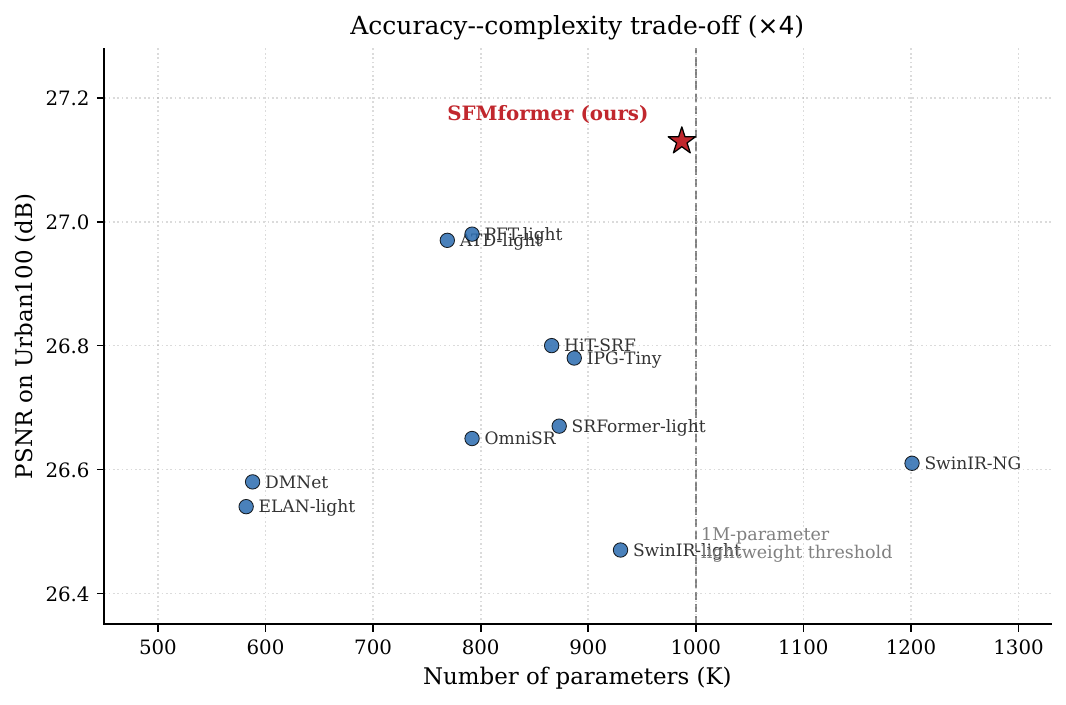}
\caption{PSNR on Urban100 versus parameter count at $\times4$. The dashed line
marks the 1M-parameter boundary conventionally used to delimit lightweight SR
models.}
\label{fig:tradeoff}
\end{figure*}

Three comparisons are worth isolating. Against HiT-SRF \cite{zhang2024hitsr},
the source of our selection-side module, the margin reaches $+0.40$~dB on
Urban100 at $\times2$. Against the two designs that place a spectral prior
elsewhere---SwinFIR-T \cite{zhang2022swinfir} at the trunk's tail and
FreqFormer \cite{dai2024freqformer} inside the attention---SFMformer leads on
every reported entry except a tie with LKMN \cite{hu2025lkmn} on Set14 PSNR at
$\times2$, with the largest margins on Manga109 ($+0.08$~dB over FreqFormer at
$\times2$ and $\times3$; $+0.18$~dB over SwinFIR-T at $\times2$). And across
datasets the improvement is systematically larger on Urban100 and Manga109
(0.15--0.31~dB over the second best) than on Set5 and BSD100 (0.04--0.10~dB),
which is the ordering our account predicts: those two benchmarks combine
regular geometric structure, which makes token selection decisive, with dense
high-frequency texture, which is what spectral modulation acts on.
Figure~\ref{fig:tradeoff} places these results on the accuracy--complexity
plane.

\begin{table*}[!htbp]
\centering
\caption{Quantitative comparison of SFMformer with state-of-the-art lightweight
methods at three upscaling factors. \best{Bold}: best; \second{underline}:
second best; bicubic is listed for reference and excluded from the ranking.
$\dagger$ denotes retraining under our configuration. LKMN
\cite{hu2025lkmn} and PromptSR \cite{liu2026promptsr} report $\times2$ results
only, and FLOPs are not available for every quoted method.}
\label{tab:sota}
\setlength{\tabcolsep}{2.5pt}
\resizebox{\textwidth}{!}{%
\begin{tabular}{llrrcccccccccc}
\toprule
\multirow{2}{*}{Method} & \multirow{2}{*}{Scale} &
\multirow{2}{*}{Params} & \multirow{2}{*}{FLOPs} &
\multicolumn{2}{c}{Set5} & \multicolumn{2}{c}{Set14} &
\multicolumn{2}{c}{BSD100} & \multicolumn{2}{c}{Urban100} &
\multicolumn{2}{c}{Manga109}\\
\cmidrule(lr){5-6}\cmidrule(lr){7-8}\cmidrule(lr){9-10}
\cmidrule(lr){11-12}\cmidrule(lr){13-14}
& & & & PSNR & SSIM & PSNR & SSIM & PSNR & SSIM & PSNR & SSIM & PSNR & SSIM\\
\midrule
Bicubic & $\times2$ & --- & --- & 33.66 & 0.9299 & 30.24 & 0.8688 & 29.56 & 0.8431 & 26.88 & 0.8403 & 30.80 & 0.9339\\
SwinIR-light \cite{liang2021swinir} & $\times2$ & 910K & 244G & 38.14 & 0.9611 & 33.86 & 0.9206 & 32.31 & 0.9012 & 32.76 & 0.9340 & 39.12 & 0.9783\\
ELAN-light \cite{zhang2022elan} & $\times2$ & 582K & 203G & 38.17 & 0.9611 & 33.94 & 0.9207 & 32.30 & 0.9012 & 32.76 & 0.9340 & 39.11 & 0.9782\\
SwinIR-NG \cite{choi2023swinirng} & $\times2$ & 1181K & 274.1G & 38.17 & 0.9612 & 33.94 & 0.9205 & 32.31 & 0.9013 & 32.78 & 0.9340 & 39.20 & 0.9781\\
OmniSR \cite{wang2023omnisr} & $\times2$ & 772K & 194.5G & 38.22 & 0.9613 & 33.98 & 0.9210 & 32.36 & 0.9020 & 33.05 & 0.9363 & 39.28 & 0.9784\\
DMNet \cite{li2025dmnet} & $\times2$ & 572K & 115.3G & 38.23 & 0.9613 & 33.95 & 0.9209 & 32.31 & 0.9015 & 32.84 & 0.9347 & 39.39 & 0.9766\\
SRFormer-light \cite{zhou2023srformer} & $\times2$ & 853K & 236G & 38.23 & 0.9615 & 33.94 & 0.9209 & 32.36 & 0.9013 & 32.91 & 0.9353 & 39.28 & 0.9785\\
HiT-SRF \cite{zhang2024hitsr} & $\times2$ & 847K & 226.5G & 38.26 & 0.9615 & 34.01 & 0.9214 & 32.37 & 0.9023 & 33.13 & 0.9372 & 39.47 & 0.9787\\
SwinFIR-T \cite{zhang2022swinfir} & $\times2$ & 872K & --- & 38.26 & 0.9616 & 34.08 & 0.9221 & 32.38 & 0.9024 & 33.14 & 0.9374 & 39.55 & 0.9790\\
IPG-Tiny \cite{tian2024ipg} & $\times2$ & 872K & 245.2G & 38.27 & 0.9616 & \best{34.24} & \best{0.9236} & 32.35 & 0.9018 & 33.04 & 0.9359 & 39.31 & 0.9786\\
ATD-light \cite{zhang2024atd} & $\times2$ & 753K & 348.6G & 38.28 & 0.9616 & 34.11 & 0.9217 & 32.39 & 0.9023 & 33.27 & 0.9376 & 39.51 & 0.9789\\
PromptSR \cite{liu2026promptsr} & $\times2$ & 764K & --- & 38.30 & 0.9617 & 34.10 & 0.9221 & 32.37 & 0.9022 & \second{33.39} & \second{0.9390} & 39.56 & 0.9790\\
FreqFormer \cite{dai2024freqformer} & $\times2$ & 870K & --- & 38.31 & 0.9616 & 34.12 & 0.9220 & 32.41 & 0.9026 & 33.25 & 0.9374 & \second{39.65} & \second{0.9792}\\
LKMN \cite{hu2025lkmn} & $\times2$ & 889K & --- & 38.32 & \second{0.9618} & \second{34.20} & 0.9223 & \second{32.43} & \second{0.9030} & 33.13 & 0.9377 & 39.54 & 0.9791\\
PFT-light$^{\dagger}$ \cite{long2025pft} & $\times2$ & 776K & 278.3G & \second{38.33} & \second{0.9618} & 34.06 & 0.9218 & 32.41 & 0.9026 & 33.36 & 0.9385 & 39.50 & 0.9790\\
\textbf{SFMformer (ours)} & $\times2$ & 970K & 315.1G & \best{38.40} & \best{0.9620} & \second{34.20} & \second{0.9227} & \best{32.45} & \best{0.9032} & \best{33.53} & \best{0.9397} & \best{39.73} & \best{0.9794}\\
\midrule
Bicubic & $\times3$ & --- & --- & 30.39 & 0.8682 & 27.55 & 0.7742 & 27.21 & 0.7385 & 24.46 & 0.7349 & 26.95 & 0.8556\\
SwinIR-light \cite{liang2021swinir} & $\times3$ & 918K & 111G & 34.62 & 0.9289 & 30.54 & 0.8463 & 29.20 & 0.8082 & 28.66 & 0.8624 & 33.98 & 0.9478\\
ELAN-light \cite{zhang2022elan} & $\times3$ & 590K & 90.1G & 34.61 & 0.9288 & 30.55 & 0.8463 & 29.21 & 0.8081 & 28.69 & 0.8624 & 34.00 & 0.9478\\
SwinIR-NG \cite{choi2023swinirng} & $\times3$ & 1190K & 114.1G & 34.64 & 0.9293 & 30.58 & 0.8471 & 29.24 & 0.8090 & 28.75 & 0.8639 & 34.22 & 0.9488\\
OmniSR \cite{wang2023omnisr} & $\times3$ & 780K & 88.4G & 34.70 & 0.9294 & 30.57 & 0.8469 & 29.28 & 0.8094 & 28.84 & 0.8656 & 34.22 & 0.9487\\
DMNet \cite{li2025dmnet} & $\times3$ & 579K & 52.0G & 34.71 & 0.9295 & 30.57 & 0.8459 & 29.26 & 0.8093 & 28.80 & 0.8640 & 34.33 & 0.9488\\
SRFormer-light \cite{zhou2023srformer} & $\times3$ & 861K & 105G & 34.67 & 0.9296 & 30.57 & 0.8469 & 29.26 & 0.8099 & 28.81 & 0.8655 & 34.19 & 0.9489\\
HiT-SRF \cite{zhang2024hitsr} & $\times3$ & 855K & 101.6G & 34.75 & 0.9300 & 30.61 & 0.8475 & 29.29 & 0.8106 & 28.99 & 0.8687 & 34.53 & 0.9502\\
SwinFIR-T \cite{zhang2022swinfir} & $\times3$ & 880K & --- & 34.75 & 0.9300 & 30.68 & \second{0.8489} & 29.30 & 0.8106 & 29.04 & 0.8697 & 34.60 & 0.9506\\
IPG-Tiny \cite{tian2024ipg} & $\times3$ & 878K & 109.0G & 34.64 & 0.9292 & 30.61 & 0.8470 & 29.26 & 0.8097 & 28.93 & 0.8666 & 34.30 & 0.9493\\
ATD-light \cite{zhang2024atd} & $\times3$ & 760K & 154.7G & 34.70 & 0.9300 & 30.68 & 0.8485 & 29.32 & 0.8109 & 29.16 & 0.8710 & 34.60 & 0.9505\\
FreqFormer \cite{dai2024freqformer} & $\times3$ & 878K & --- & \second{34.86} & \second{0.9307} & \second{30.71} & 0.8488 & \second{29.35} & 0.8116 & 29.15 & 0.8710 & \second{34.80} & \second{0.9513}\\
PFT-light$^{\dagger}$ \cite{long2025pft} & $\times3$ & 783K & 123.5G & 34.78 & 0.9303 & 30.70 & 0.8481 & 29.32 & \second{0.8122} & \second{29.22} & \second{0.8728} & 34.57 & 0.9505\\
\textbf{SFMformer (ours)} & $\times3$ & 977K & 140.6G & \best{34.88} & \best{0.9311} & \best{30.79} & \best{0.8500} & \best{29.38} & \best{0.8125} & \best{29.37} & \best{0.8744} & \best{34.88} & \best{0.9516}\\
\midrule
Bicubic & $\times4$ & --- & --- & 28.42 & 0.8104 & 26.00 & 0.7027 & 25.96 & 0.6675 & 23.14 & 0.6577 & 24.89 & 0.7866\\
SwinIR-light \cite{liang2021swinir} & $\times4$ & 930K & 63.6G & 32.44 & 0.8976 & 28.77 & 0.7858 & 27.69 & 0.7406 & 26.47 & 0.7980 & 30.92 & 0.9151\\
ELAN-light \cite{zhang2022elan} & $\times4$ & 582K & 54.1G & 32.43 & 0.8975 & 28.78 & 0.7858 & 27.69 & 0.7406 & 26.54 & 0.7982 & 30.92 & 0.9150\\
SwinIR-NG \cite{choi2023swinirng} & $\times4$ & 1201K & 63.0G & 32.44 & 0.8980 & 28.83 & 0.7870 & 27.73 & 0.7418 & 26.61 & 0.8010 & 31.09 & 0.9161\\
OmniSR \cite{wang2023omnisr} & $\times4$ & 792K & 50.9G & 32.49 & 0.8988 & 28.78 & 0.7859 & 27.71 & 0.7415 & 26.65 & 0.8018 & 31.02 & 0.9151\\
DMNet \cite{li2025dmnet} & $\times4$ & 588K & 29.7G & 32.51 & 0.8987 & 28.84 & 0.7866 & 27.73 & 0.7410 & 26.58 & 0.7991 & 31.14 & 0.9150\\
IPG-Tiny \cite{tian2024ipg} & $\times4$ & 887K & 61.3G & 32.51 & 0.8987 & 28.85 & 0.7873 & 27.73 & 0.7418 & 26.78 & 0.8050 & 31.22 & 0.9176\\
SRFormer-light \cite{zhou2023srformer} & $\times4$ & 873K & 62.8G & 32.51 & 0.8988 & 28.82 & 0.7872 & 27.73 & 0.7422 & 26.67 & 0.8032 & 31.17 & 0.9165\\
HiT-SRF \cite{zhang2024hitsr} & $\times4$ & 866K & 58.0G & 32.55 & 0.8999 & 28.87 & 0.7880 & 27.75 & 0.7432 & 26.80 & 0.8069 & 31.26 & 0.9171\\
ATD-light \cite{zhang2024atd} & $\times4$ & 769K & 87.1G & 32.62 & 0.8997 & 28.87 & 0.7884 & 27.77 & 0.7439 & 26.97 & 0.8107 & 31.47 & 0.9198\\
SwinFIR-T \cite{zhang2022swinfir} & $\times4$ & 891K & --- & 32.62 & 0.9002 & \second{28.95} & \second{0.7898} & \second{27.79} & 0.7440 & 26.85 & 0.8088 & 31.50 & 0.9199\\
PFT-light$^{\dagger}$ \cite{long2025pft} & $\times4$ & 792K & 69.6G & 32.55 & 0.8992 & 28.91 & 0.7887 & 27.77 & 0.7435 & \second{26.98} & \second{0.8120} & 31.45 & 0.9192\\
FreqFormer \cite{dai2024freqformer} & $\times4$ & 889K & --- & \second{32.69} & \second{0.9007} & \second{28.95} & \second{0.7898} & \second{27.79} & \second{0.7444} & 26.84 & 0.8093 & \second{31.59} & \second{0.9201}\\
\textbf{SFMformer (ours)} & $\times4$ & 987K & 79.8G & \best{32.70} & \best{0.9009} & \best{28.98} & \best{0.7900} & \best{27.82} & \best{0.7450} & \best{27.13} & \best{0.8158} & \best{31.65} & \best{0.9212}\\
\bottomrule
\end{tabular}}
\end{table*}

\subsection{Effect of batch size}
\label{sec:exp_bs}

To examine the influence of batch size on model performance, we retrained
SFMformer and PFT-light \cite{long2025pft} with a batch size of 32; results are
listed in Table~\ref{tab:bs32}. All other training settings (patch size,
optimiser, iteration count and so on) are identical to
Table~\ref{tab:sota}---only the batch size differs. The remaining reference
figures are still quoted from the original papers, so this table serves mainly
to observe the effect of batch size on SFMformer and its baseline, and is
supplementary to Table~\ref{tab:sota}.

\begin{table*}[!htbp]
\centering
\caption{Results at a batch size of 32. PFT-light$^{\dagger}$ was retrained by
us under settings identical to SFMformer; the PFT-light rows without $\dagger$
quote the original paper \cite{long2025pft}. \best{Bold}: best;
\second{underline}: second best.}
\label{tab:bs32}
\setlength{\tabcolsep}{3pt}
\small
\resizebox{\textwidth}{!}{%
\begin{tabular}{llrcccccccccc}
\toprule
\multirow{2}{*}{Method} & \multirow{2}{*}{Scale} & \multirow{2}{*}{Params} &
\multicolumn{2}{c}{Set5} & \multicolumn{2}{c}{Set14} &
\multicolumn{2}{c}{BSD100} & \multicolumn{2}{c}{Urban100} &
\multicolumn{2}{c}{Manga109}\\
\cmidrule(lr){4-5}\cmidrule(lr){6-7}\cmidrule(lr){8-9}
\cmidrule(lr){10-11}\cmidrule(lr){12-13}
& & & PSNR & SSIM & PSNR & SSIM & PSNR & SSIM & PSNR & SSIM & PSNR & SSIM\\
\midrule
PFT-light \cite{long2025pft} & $\times2$ & 776K & \second{38.36} & \second{0.9620} & \second{34.19} & \second{0.9232} & 32.43 & \second{0.9030} & \second{33.67} & \second{0.9411} & 39.55 & 0.9792\\
PFT-light$^{\dagger}$ & $\times2$ & 776K & 38.35 & \second{0.9620} & \second{34.19} & 0.9227 & \second{32.44} & \second{0.9030} & 33.61 & 0.9406 & \second{39.64} & \second{0.9793}\\
\textbf{SFMformer (ours)} & $\times2$ & 970K & \best{38.43} & \best{0.9622} & \best{34.28} & \best{0.9235} & \best{32.48} & \best{0.9036} & \best{33.73} & \best{0.9414} & \best{39.82} & \best{0.9796}\\
\midrule
PFT-light \cite{long2025pft} & $\times3$ & 783K & \second{34.82} & 0.9305 & \second{30.75} & \second{0.8493} & 29.33 & 0.8116 & \best{29.46} & \best{0.8763} & 34.61 & 0.9509\\
PFT-light$^{\dagger}$ & $\times3$ & 783K & \second{34.82} & \second{0.9306} & \second{30.75} & 0.8490 & \second{29.34} & \second{0.8118} & 29.37 & \second{0.8752} & \second{34.68} & \second{0.9511}\\
\textbf{SFMformer (ours)} & $\times3$ & 977K & \best{34.89} & \best{0.9312} & \best{30.79} & \best{0.8498} & \best{29.39} & \best{0.8126} & \second{29.43} & \second{0.8752} & \best{34.88} & \best{0.9517}\\
\midrule
PFT-light \cite{long2025pft} & $\times4$ & 792K & \second{32.63} & \second{0.9005} & 28.92 & 0.7891 & 27.78 & 0.7442 & \best{27.20} & \best{0.8171} & 31.51 & \second{0.9204}\\
PFT-light$^{\dagger}$ & $\times4$ & 792K & 32.62 & 0.9000 & \second{28.96} & \second{0.7898} & \second{27.79} & \second{0.7445} & \second{27.11} & \second{0.8154} & \second{31.57} & \best{0.9207}\\
\textbf{SFMformer (ours)} & $\times4$ & 987K & \best{32.68} & \best{0.9012} & \best{29.00} & \best{0.7901} & \best{27.83} & \best{0.7448} & 27.08 & 0.8128 & \best{31.66} & \best{0.9207}\\
\bottomrule
\end{tabular}}
\end{table*}

With the batch size raised to 32, SFMformer outperforms the retrained
baseline PFT-light$^{\dagger}$ on Set5, Set14, BSD100 and Manga109 at all three
scales, confirming that the proposed modules remain effective at a larger batch
size. On Set14 at $\times2$ the PSNR of SFMformer now exceeds that of IPG-Tiny
\cite{tian2024ipg} and the SSIM gap narrows.

Urban100 is the exception and deserves explicit comment rather than
explanation away. At $\times4$, SFMformer (27.08~dB) falls below both the
retrained baseline (27.11~dB) and the published PFT-light figure (27.20~dB),
reversing the $+0.15$~dB lead observed at a batch size of 8; at $\times3$ it is
0.03~dB below the published figure while remaining 0.06~dB above the retrained
one. Three observations bound the interpretation. First, the effect is confined
to one benchmark: SFMformer leads PFT-light$^{\dagger}$ on the other four at
every scale under an identical configuration. Second, all figures here come
from a single training run per configuration, and the differences at issue are a
few hundredths of a decibel; we did not have the compute budget to average over
random seeds, so we can claim neither that the gap lies outside run-to-run
variation nor that it lies inside it. Third, Urban100 rewards the exploitation
of long-range self-similarity, which SFMformer inherits from PFT unchanged, so
it is also the benchmark on which the proposed modules have least room to act---
consistent with Table~\ref{tab:interaction}, where Urban100 at $\times4$ is one
of the pairs on which the two modules fail to compound. We regard this as an
open point, and a multi-seed study of the $\times4$ setting is the natural next
step.

\subsection{Ablation study}
\label{sec:exp_ablation}

To verify the specific contribution of the DFE \cite{zhang2024hitsr} and WMA
\cite{li2025dmnet} modules, we conduct an ablation study using PFT-light
\cite{long2025pft} as the baseline and adding DFE alone, WMA alone, or both
(the complete SFMformer), giving four variants. All ablation experiments use
exactly the training configuration of Section~\ref{sec:exp_impl} (batch size 8,
patch size, optimiser, learning-rate schedule and iteration count included) and
are trained from random initialisation on DF2K. Results are given in
Table~\ref{tab:ablation}.

\begin{table*}[!htbp]
\centering
\caption{Ablation study of the DFE and WMA modules at all three upscaling
factors.}
\label{tab:ablation}
\setlength{\tabcolsep}{2.5pt}
\resizebox{\textwidth}{!}{%
\begin{tabular}{lccrrcccccccccc}
\toprule
\multirow{2}{*}{Method} & \multirow{2}{*}{DFE} & \multirow{2}{*}{WMA} &
\multirow{2}{*}{Scale} & \multirow{2}{*}{Params} &
\multicolumn{2}{c}{Set5} & \multicolumn{2}{c}{Set14} &
\multicolumn{2}{c}{BSD100} & \multicolumn{2}{c}{Urban100} &
\multicolumn{2}{c}{Manga109}\\
\cmidrule(lr){6-7}\cmidrule(lr){8-9}\cmidrule(lr){10-11}
\cmidrule(lr){12-13}\cmidrule(lr){14-15}
& & & & & PSNR & SSIM & PSNR & SSIM & PSNR & SSIM & PSNR & SSIM & PSNR & SSIM\\
\midrule
PFT-light$^{\dagger}$ (baseline) & & & $\times2$ & 776K & 38.33 & 0.9618 & 34.06 & 0.9218 & 32.41 & 0.9026 & 33.36 & 0.9385 & 39.50 & 0.9790\\
\quad + DFE & \checkmark & & $\times2$ & 890K & 38.35 & \best{0.9620} & 34.15 & 0.9221 & 32.43 & 0.9029 & 33.47 & \best{0.9398} & 39.60 & 0.9791\\
\quad + WMA & & \checkmark & $\times2$ & 856K & 38.36 & 0.9619 & 34.05 & 0.9220 & 32.42 & 0.9028 & 33.37 & 0.9382 & 39.65 & 0.9793\\
\quad + DFE + WMA (SFMformer) & \checkmark & \checkmark & $\times2$ & 970K & \best{38.40} & \best{0.9620} & \best{34.20} & \best{0.9227} & \best{32.45} & \best{0.9032} & \best{33.53} & 0.9397 & \best{39.73} & \best{0.9794}\\
\midrule
PFT-light$^{\dagger}$ (baseline) & & & $\times3$ & 783K & 34.78 & 0.9303 & 30.70 & 0.8481 & 29.32 & 0.8122 & 29.22 & 0.8728 & 34.57 & 0.9505\\
\quad + DFE & \checkmark & & $\times3$ & 897K & 34.81 & 0.9307 & 30.73 & 0.8491 & 29.35 & 0.8117 & 29.33 & \best{0.8748} & 34.72 & 0.9511\\
\quad + WMA & & \checkmark & $\times3$ & 863K & 34.83 & 0.9307 & 30.73 & 0.8488 & 29.34 & 0.8115 & 29.25 & 0.8719 & 34.74 & 0.9509\\
\quad + DFE + WMA (SFMformer) & \checkmark & \checkmark & $\times3$ & 977K & \best{34.88} & \best{0.9311} & \best{30.79} & \best{0.8500} & \best{29.38} & \best{0.8125} & \best{29.37} & 0.8744 & \best{34.88} & \best{0.9516}\\
\midrule
PFT-light$^{\dagger}$ (baseline) & & & $\times4$ & 792K & 32.55 & 0.8992 & 28.91 & 0.7887 & 27.77 & 0.7435 & 26.98 & 0.8120 & 31.45 & 0.9192\\
\quad + DFE & \checkmark & & $\times4$ & 907K & 32.55 & 0.8998 & 28.95 & 0.7899 & 27.80 & 0.7446 & 27.09 & 0.8151 & 31.57 & 0.9209\\
\quad + WMA & & \checkmark & $\times4$ & 873K & 32.61 & 0.9000 & 28.96 & 0.7894 & 27.81 & 0.7441 & 27.03 & 0.8109 & 31.64 & 0.9205\\
\quad + DFE + WMA (SFMformer) & \checkmark & \checkmark & $\times4$ & 987K & \best{32.70} & \best{0.9009} & \best{28.98} & \best{0.7900} & \best{27.82} & \best{0.7450} & \best{27.13} & \best{0.8158} & \best{31.65} & \best{0.9212}\\
\bottomrule
\end{tabular}}
\end{table*}

Each module helps on its own, but in different places. DFE gives a small,
uniform improvement on every dataset---largest on Urban100 ($+0.11$~dB at
$\times2$), whose regular geometry rewards sharper spatial discrimination---for
$+114$K parameters ($+14.7\%$), behaving as a general-purpose enhancement. WMA
instead concentrates its benefit on Manga109 ($+0.15$ to $+0.19$~dB across
scales), whose line art puts most of its energy in the high-frequency sub-bands
WMA acts on, and is neutral or marginally negative elsewhere, for $+80$K
parameters ($+10.3\%$). Enabled together they give the best result on almost
every entry.

\subsubsection{When do the two modules compound?}
\label{sec:exp_interaction}

With both modules enabled---the complete SFMformer---the model attains the best
performance on almost every dataset. The more informative question is whether
the joint gain is merely the sum of the parts. Writing $\Delta_{\mathrm{DFE}}$
and $\Delta_{\mathrm{WMA}}$ for the PSNR gain of each module in isolation over
the baseline and $\Delta_{\mathrm{both}}$ for the gain with both enabled, we
define the interaction term

\begin{equation}
\varepsilon = \Delta_{\mathrm{both}}
              - (\Delta_{\mathrm{DFE}} + \Delta_{\mathrm{WMA}}),
\label{eq:interaction}
\end{equation}

which is positive when the modules compound, zero when they act independently
and negative when their gains overlap. Table~\ref{tab:interaction} reports
$\varepsilon$ for all fifteen benchmark--scale pairs.

The result is not uniform, and the structure of the exceptions is what makes it
interpretable. The interaction is positive on nine of the fifteen pairs,
reaching $+0.06$~dB on Set14 at $\times2$ and $+0.09$~dB on Set5 at $\times4$,
where the measured joint gain is roughly one and a half to two times what
independence would predict. It is negative on all three Manga109 entries
($-0.02$, $-0.01$, $-0.11$~dB) and on three further entries at $\times4$.

The two regimes are separated cleanly by a single quantity: the gain of the
\emph{weaker} of the two interventions,
$\min(\Delta_{\mathrm{DFE}},\Delta_{\mathrm{WMA}})$, which correlates with
$\varepsilon$ at $r=-0.72$---a stronger association than either intervention's
own gain ($r=-0.40$ for DFE, $r=-0.67$ for WMA) or their asymmetry
($r=+0.16$). Thresholding it makes the split almost categorical: of the ten
pairs where the weaker intervention contributes at most $0.03$~dB, nine have
$\varepsilon>0$; of the five where it contributes more, none does.

Under the account of Section~\ref{sec:intro_contrib} this is what one expects.
Where selection is the binding constraint and little is available from spectral
correction---Urban100, whose regular geometry makes the ranking of tokens
decisive but whose energy is not concentrated in any one sub-band---the two
interventions relieve different constraints and their gains compound. Where
both interventions end up improving the same content they cannot compound:
Manga109 is almost entirely high-contrast line art, so the spatial branch
sharpens exactly the edge features whose sub-bands the spectral branch is
amplifying, and the two claim overlapping portions of the same headroom. That
five of the six negative entries occur at $\times4$ fits the same reading, since
the more severe degradation leaves less high-frequency information for either
intervention to recover and therefore less room for the two to divide.

\subsubsection*{What this evidence does and does not establish}

The pattern above is consistent with the selection/aggregation account, but it
does not by itself establish it, and we want to be explicit about the
alternative. A generic diminishing-returns model---in which any two improvements
to the same network compound sub-linearly once both are large, regardless of
what they do---predicts a negative association between $\varepsilon$ and
$\min(\Delta_{\mathrm{DFE}},\Delta_{\mathrm{WMA}})$ without any appeal to
distinct stages. Our measurements do not discriminate between these
two explanations, because both predict the sign pattern we observe.

Two experiments would separate them, and we identify them as the natural
continuation of this work rather than claiming their outcome. The first is
diagnostic and requires no additional training: extracting the top-$k$ index
sets of the trained baseline and of the baseline with spatial enhancement added,
and measuring both their overlap and the fraction of retained tokens falling on
high-gradient regions. The selection account predicts that spatial enhancement
measurably changes \emph{which} tokens survive, and changes them toward
structural content; a generic feature-quality account predicts the selected sets
remain largely the same while the values attached to them improve. The second is
a transfer test: applying the same spatial enhancement to a dense-attention
backbone such as SwinIR-light and comparing the gain to the one measured here.
If the selection stage is what the enhancement exploits, its benefit should be
substantially smaller where no selection stage exists.

Two further observations bear on the design as a whole. The joint configuration
is never harmful---$\Delta_{\mathrm{both}}$ is the largest of the three gains on
fourteen of fifteen pairs---so the pairing is worth having even on the
benchmarks where it does not compound. And a positive $\varepsilon$ is difficult
to obtain from two modules acting at the same point in the pipeline, which is
weak but real evidence that the ordering
DFE~$\rightarrow$~PFA~$\rightarrow$~WMA is doing structural work rather than
simply adding capacity.

\begin{table}[!htbp]
\centering
\caption{Module interaction across all benchmark--scale pairs. $\Delta$ values
are PSNR gains in dB over the PFT-light$^{\dagger}$ baseline;
$\varepsilon$ is the interaction term of Eq.~\eqref{eq:interaction}. Positive
$\varepsilon$ means the modules compound; negative means their gains overlap.}
\label{tab:interaction}
\small
\begin{tabular}{llrrrrr}
\toprule
Dataset & Scale & $\Delta_{\mathrm{DFE}}$ & $\Delta_{\mathrm{WMA}}$ &
Sum & $\Delta_{\mathrm{both}}$ & $\varepsilon$\\
\midrule
Set5     & $\times2$ & $+0.02$ & $+0.03$ & $+0.05$ & $+0.07$ & $+0.02$\\
Set14    & $\times2$ & $+0.09$ & $-0.01$ & $+0.08$ & $+0.14$ & $\best{+0.06}$\\
BSD100   & $\times2$ & $+0.02$ & $+0.01$ & $+0.03$ & $+0.04$ & $+0.01$\\
Urban100 & $\times2$ & $+0.11$ & $+0.01$ & $+0.12$ & $+0.17$ & $+0.05$\\
Manga109 & $\times2$ & $+0.10$ & $+0.15$ & $+0.25$ & $+0.23$ & $-0.02$\\
\midrule
Set5     & $\times3$ & $+0.03$ & $+0.05$ & $+0.08$ & $+0.10$ & $+0.02$\\
Set14    & $\times3$ & $+0.03$ & $+0.03$ & $+0.06$ & $+0.09$ & $+0.03$\\
BSD100   & $\times3$ & $+0.03$ & $+0.02$ & $+0.05$ & $+0.06$ & $+0.01$\\
Urban100 & $\times3$ & $+0.11$ & $+0.03$ & $+0.14$ & $+0.15$ & $+0.01$\\
Manga109 & $\times3$ & $+0.15$ & $+0.17$ & $+0.32$ & $+0.31$ & $-0.01$\\
\midrule
Set5     & $\times4$ & $+0.00$ & $+0.06$ & $+0.06$ & $+0.15$ & $\best{+0.09}$\\
Set14    & $\times4$ & $+0.04$ & $+0.05$ & $+0.09$ & $+0.07$ & $-0.02$\\
BSD100   & $\times4$ & $+0.03$ & $+0.04$ & $+0.07$ & $+0.05$ & $-0.02$\\
Urban100 & $\times4$ & $+0.11$ & $+0.05$ & $+0.16$ & $+0.15$ & $-0.01$\\
Manga109 & $\times4$ & $+0.12$ & $+0.19$ & $+0.31$ & $+0.20$ & $-0.11$\\
\bottomrule
\end{tabular}
\end{table}

\subsection{Qualitative comparison}
\label{sec:exp_visual}

\begin{figure*}[t]
\centering
\includegraphics[width=\textwidth]{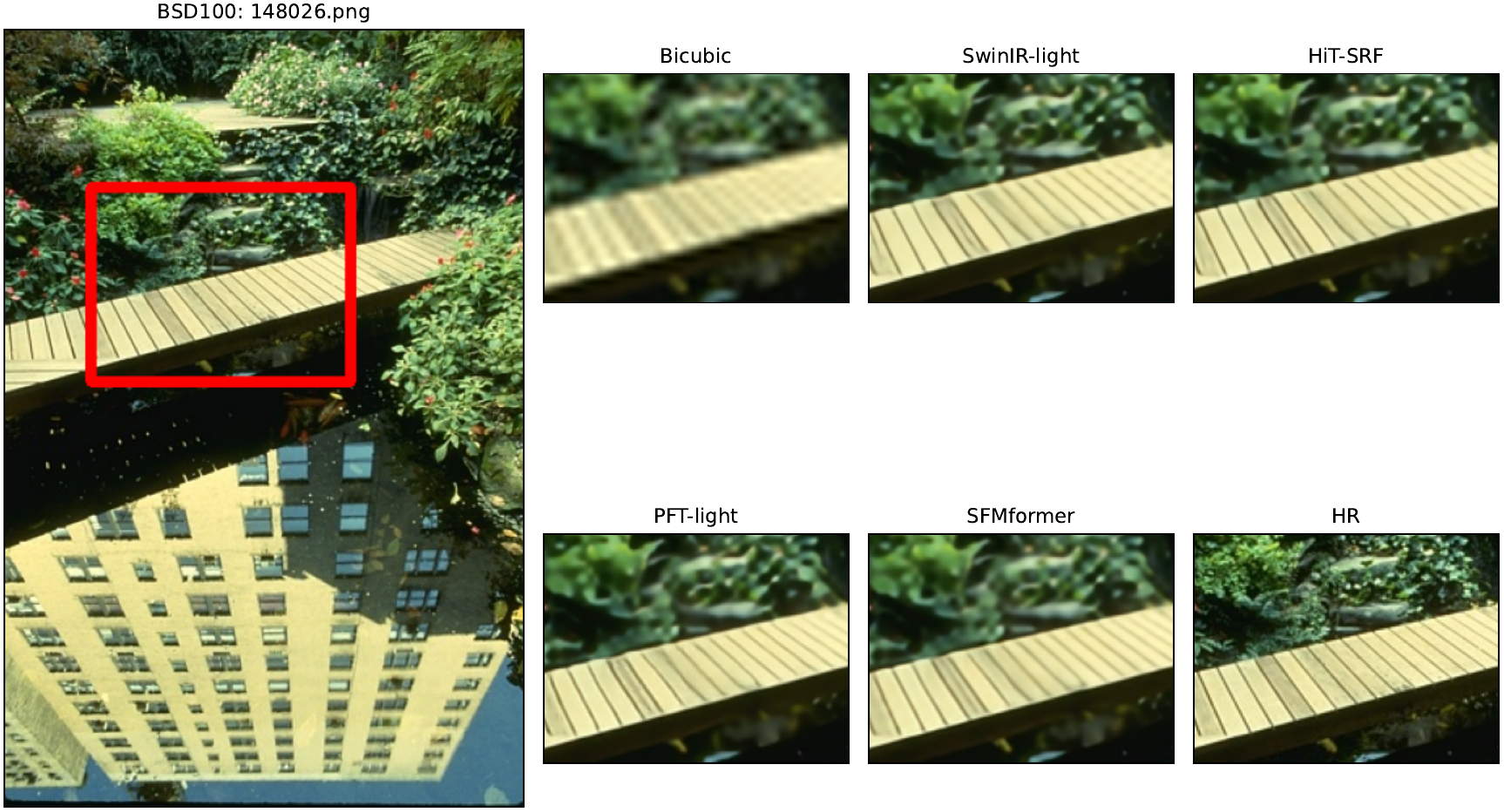}
\caption{Visual comparison on image 148026 from BSD100 at $\times4$.}
\label{fig:vis_bsd}
\end{figure*}

\begin{figure*}[t]
\centering
\includegraphics[width=\textwidth]{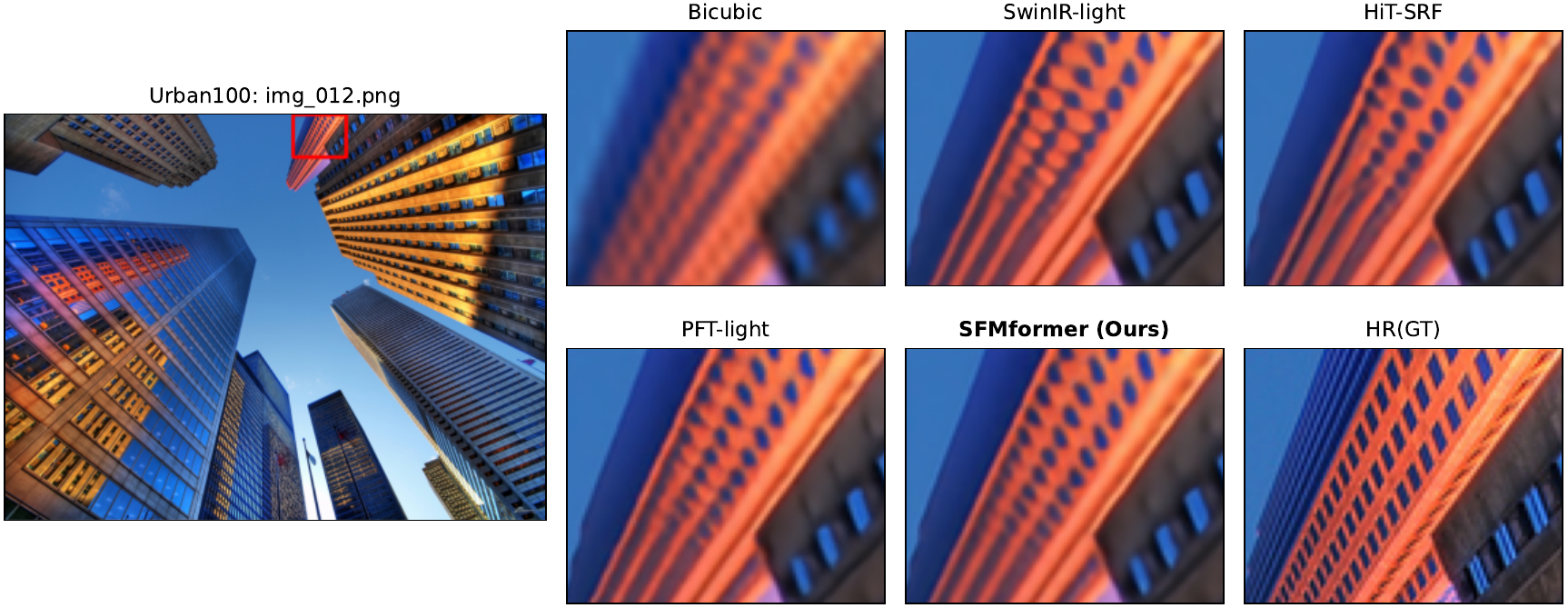}
\caption{Visual comparison on the 12th image of Urban100 at $\times4$.}
\label{fig:vis_u012}
\end{figure*}

Objective metrics do not fully capture perceived reconstruction quality, so
Figs.~\ref{fig:vis_bsd} and \ref{fig:vis_u012} compare SFMformer at $\times4$
against bicubic interpolation, SwinIR-light \cite{liang2021swinir}, HiT-SRF
\cite{zhang2024hitsr} and the baseline PFT-light \cite{long2025pft}, with the HR
image as reference.

\begin{figure*}[t]
\centering
\includegraphics[width=\textwidth]{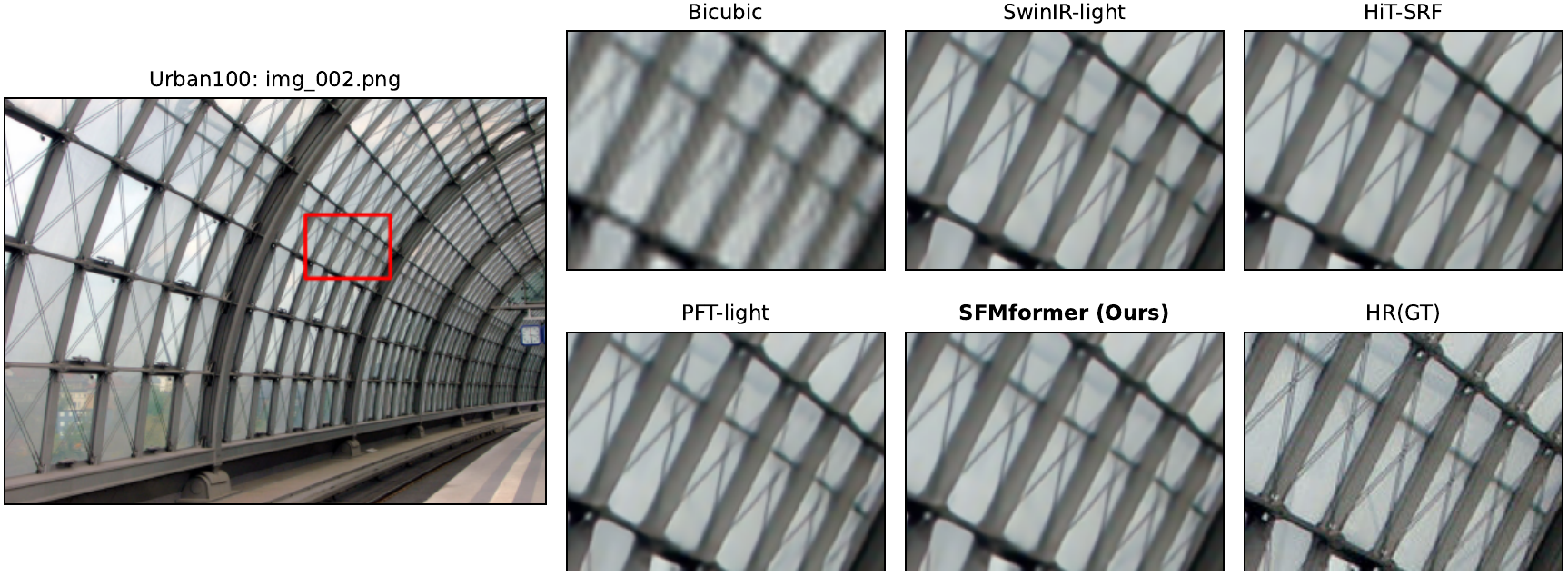}
\caption{Visual comparison on the 2nd image of Urban100 at $\times4$. At the
intersections of the X-shaped steel framing, PFT-light and HiT-SRF fade near the
horizontal members where SFMformer renders the framing distinctly.}
\label{fig:vis_u002}
\end{figure*}

\begin{figure*}[t]
\centering
\includegraphics[width=\textwidth]{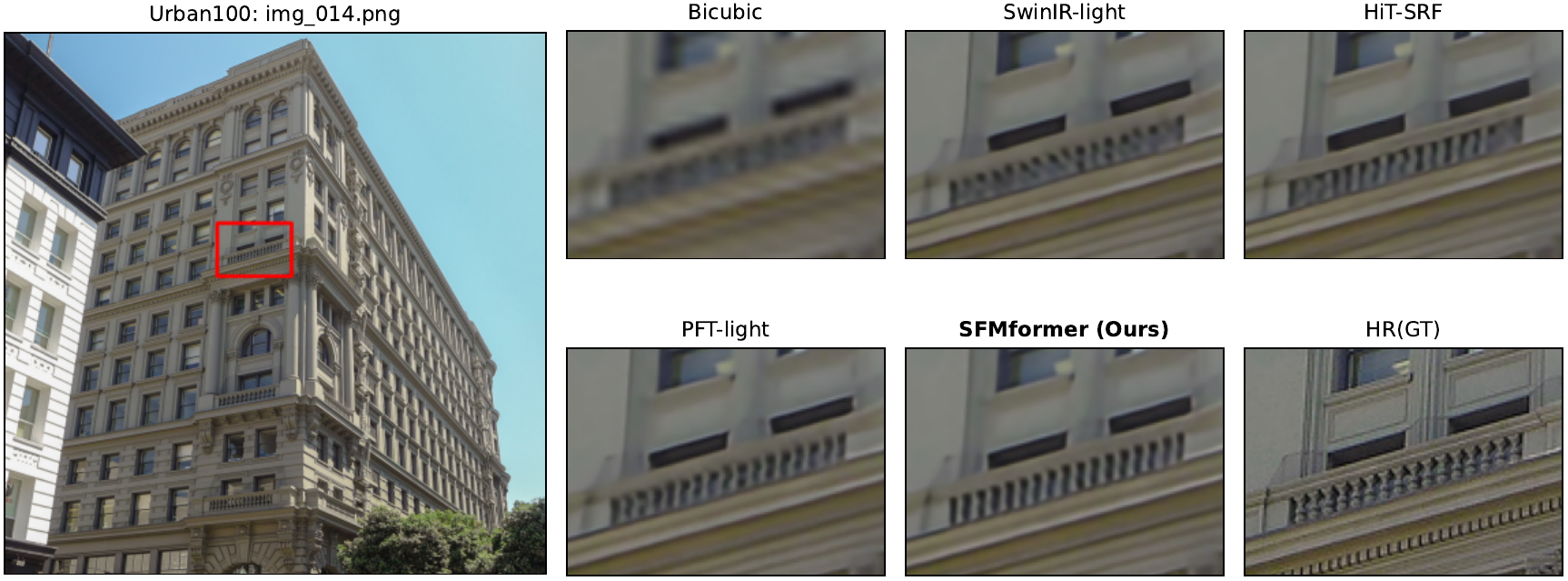}
\caption{Visual comparison on the 14th image of Urban100 at $\times4$.
SFMformer recovers the structural shape of the railing almost completely, where
the other models produce an incomplete structure or outright blur.}
\label{fig:vis_u014}
\end{figure*}

\begin{figure*}[t]
\centering
\includegraphics[width=\textwidth]{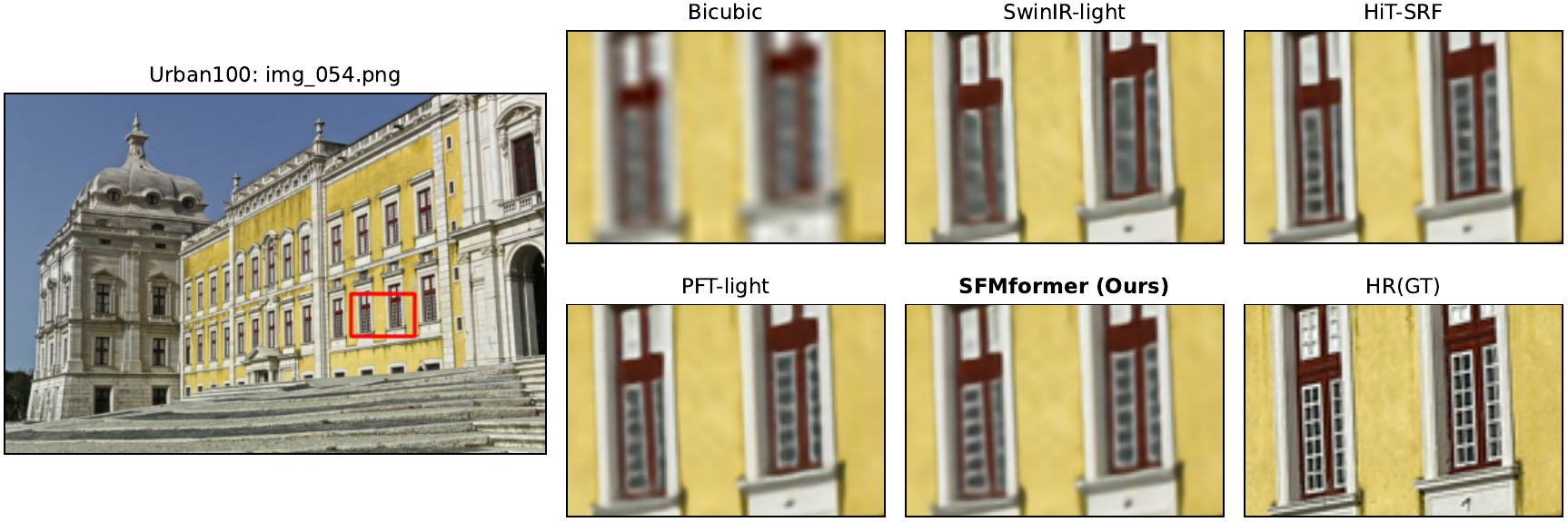}
\caption{Visual comparison on the 54th image of Urban100 at $\times4$. No model
recovers the window grid correctly; HiT-SRF cannot separate the panes, PFT-light
separates them but skews some, and SFMformer keeps almost all of them square.}
\label{fig:vis_u054}
\end{figure*}

The differences are consistent with the quantitative pattern. On the boardwalk
of Fig.~\ref{fig:vis_bsd} the plank gaps SFMformer reconstructs are straighter,
where the other models introduce visible curvature. The window grids of
Fig.~\ref{fig:vis_u012} are harder: no model recovers the structure completely,
but PFT-light leaves pronounced black smearing along the pane edges that
SFMformer avoids while separating individual panes more clearly. Figs.~\ref{fig:vis_u002}--\ref{fig:vis_u054} show the same tendency on further
Urban100 samples: railings and steel framing retain their structural shape
rather than fading or blurring at intersections, which is what one expects if
selection-side enhancement is helping the attention keep hold of the tokens
carrying regular geometry.

\subsection{Deployment on an edge device}
\label{sec:exp_edge}

\subsubsection{Deployment and application interface}

\begin{figure*}[t]
\centering
\includegraphics[width=\textwidth]{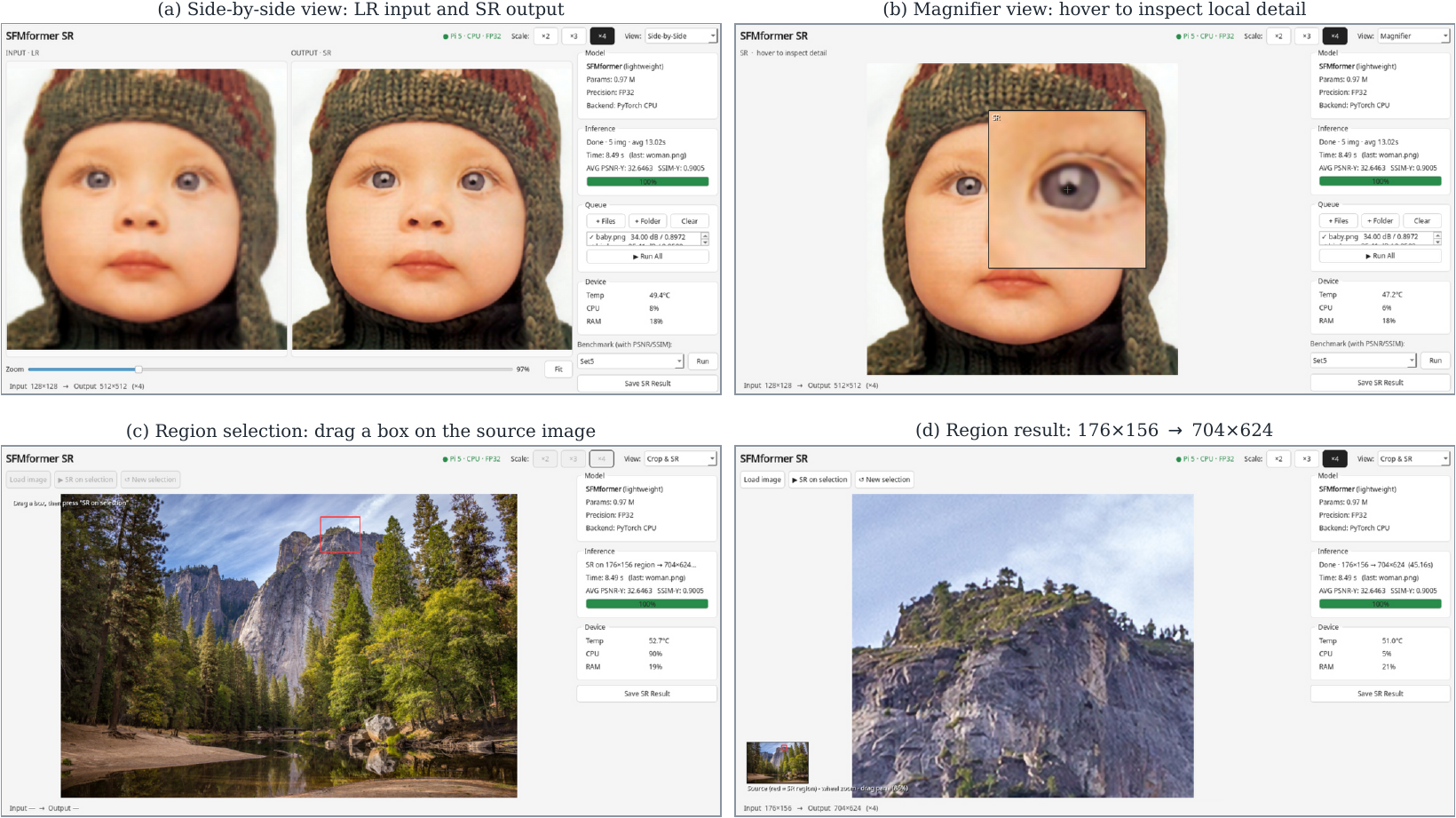}
\caption{The interactive inference interface on the Raspberry~Pi~5.
(a) Side-by-side view. (b) Magnifier view. (c) Region selection.
(d) The corresponding region result.}
\label{fig:gui}
\end{figure*}

To verify that the model is usable outside a desktop or cloud environment, we
deployed it to a Raspberry~Pi~5 (Arm Cortex-A76 quad-core @ 2.4~GHz, 16~GB
LPDDR4X, Raspberry~Pi~OS, PyTorch 2.11.0 CPU build) and built the interactive
interface of Fig.~\ref{fig:gui}. It integrates inference, magnifier and
side-by-side comparison, batch benchmarking over a test set, and device
monitoring, so evaluation can be completed on the deployment platform itself.
Because inference cost grows sharply with output size, the interface also lets
the user drag a rectangle over a region and reconstruct only that---the
mechanism whose latency implications Section~\ref{sec:exp_scenarios} examines.

The quality figures shown on screen come from the application's own metric
routine rather than the evaluation script used for Table~\ref{tab:sota}, and
differ by a few hundredths of a decibel (32.65~dB against 32.68~dB on Set5 at
$\times4$) owing to border handling and colour conversion; all figures reported
elsewhere in this paper use the standard protocol.

\subsubsection{Inference time}

Because inference runs on the CPU alone, without GPU acceleration, per-image
inference is comparatively slow. We measured inference time on the five test
sets at $\times4$; the average time per image is reported in
Table~\ref{tab:time}.

\begin{table}[!htbp]
\centering
\caption{Average per-image inference time at $\times4$ on the Raspberry Pi 5
(CPU only).}
\label{tab:time}
\small
\begin{tabular}{lc}
\toprule
Dataset & Average time per image (s)\\
\midrule
Set5 \cite{bevilacqua2012set5} & 12.85\\
Set14 \cite{zeyde2010set14} & 24.29\\
BSD100 \cite{martin2001bsd} & 17.36\\
Urban100 \cite{huang2015urban100} & 97.42\\
Manga109 \cite{matsui2017manga109} & 103.88\\
\bottomrule
\end{tabular}
\end{table}

The measured times clearly reflect image size: the larger the image, the longer
inference takes. This is most evident on Urban100 and Manga109. Urban100 images
are upscaled from roughly $256\times192$ to $1024\times768$ at $\times4$, while
Manga109 images go from about $206\times292$ to $824\times1168$, and the timings
track these dimensions directly.

\subsubsection{Which deployment patterns the measured latency supports}
\label{sec:exp_scenarios}

The timings in Table~\ref{tab:time} are informative about what this model can
and cannot be used for on CPU-only hardware, and we state the boundary plainly
rather than leaving it implied.

\emph{Interactive inspection is practical, provided the region is bounded.}
Reconstructing a full Manga109 page takes 104~s, which no operator will wait
for. But the interface of Section~\ref{sec:exp_edge} does not require it: the
$176\times156$ region selected in Fig.~\ref{fig:gui}(c) reconstructs to
$704\times624$ in 45~s, and the $128\times128$ inputs of Set5 in 13~s. Because
inference cost scales with pixel count rather than with the size of the source
image, the operator controls the latency directly by choosing how much to
enlarge. This is the pattern the region-of-interest mode exists to serve, and it
is where a sub-1M model earns its keep: the same workflow with a 12M-parameter
model would be roughly an order of magnitude slower on the same hardware and
would leave the interactive regime entirely.

\emph{Batch and archival processing is comfortable.} At 13--104~s per image
depending on resolution, a Raspberry~Pi~5 processes on the order of a thousand
images per day unattended, at a power draw of a few watts. For archival
restoration or overnight processing of a day's captures this is adequate, and
the relevant constraint becomes storage rather than compute.

\emph{Real-time video is out of reach on this hardware, and we do not claim
otherwise.} Even the smallest test images are three orders of magnitude away
from a 30~fps budget. Reaching video rates would require quantisation, an NPU or
GPU backend, or both; the FP32 PyTorch CPU path measured here is a portability
baseline, not an optimised deployment. What the measurements do establish is
that the architecture fits within the memory and thermal envelope of a
credit-card computer---peak RAM stayed at 21\% of 16~GB and the SoC settled at
$51$--$53^{\circ}$C under sustained load, well inside the throttling
threshold---so the remaining gap is an engineering one rather than a
question of whether the model fits.

\subsection{Limitations}
\label{sec:exp_limitations}

Four limitations deserve mention. First, constrained by the lightweight
parameter budget and by an optimisation objective that is distortion-oriented in
both the pixel and frequency domains, SFMformer still tends to produce somewhat
smooth output in regions of complex texture. We report only fidelity metrics
(PSNR/SSIM); no perceptual evaluation is included, whether full-reference such
as LPIPS \cite{zhang2018lpips} or no-reference such as MUSIQ \cite{ke2021musiq}
and CLIP-IQA \cite{wang2023clipiqa}. The perceptual standing of the method
relative to the compared baselines is therefore not established here, and since
the wavelet branch is motivated by high-frequency detail, this is the evaluation
gap most worth closing. Second, the proposed modules cost 15--25\% more parameters and
FLOPs than the baseline for gains of 0.1--0.3~dB; we report FLOPs and CPU
inference time on an edge device, but not GPU-side latency or throughput, so the
practical efficiency trade-off on accelerators is not quantified. Third, all
results come from a single training run per configuration, which limits what can
be concluded from small differences---see the discussion of Urban100 at
$\times4$ in Section~\ref{sec:exp_bs}. Fourth, training and testing are both
based on synthetic bicubic-downsampled pairs, which differ from the complex
degradations encountered in the real world, such as unknown blur kernels, sensor
noise and compression artefacts. The first and last points are addressed in
Section~\ref{sec:conclusion} as directions for future work.

\section{Conclusion and future work}
\label{sec:conclusion}

This paper began from an observation about sparse attention rather than about
super-resolution. Because a top-$k$ operator selects before it aggregates, and
because progressive focusing propagates that selection forward so a discarded
token cannot return, a sparse attention layer offers two separable targets for
improvement where a dense one offers a single target. We tested this by filling
both positions with deliberately ordinary components---dual-branch spatial
enhancement \cite{zhang2024hitsr} on the input of the projection and
wavelet-domain modulation \cite{li2025dmnet} on the output---around the
progressive focused attention of PFT \cite{long2025pft}.

The measurements support the prediction without settling it. The two compound
on nine of fifteen benchmark--scale pairs and overlap on six, and which occurs
is predicted by how much the weaker intervention achieves alone ($r=-0.72$). As
Section~\ref{sec:exp_interaction} notes, a generic diminishing-returns model
predicts the same sign pattern; separating the two requires the
selection-overlap diagnostic or the transfer test to a dense backbone, neither
attempted here. What the results do establish is that the paired configuration
is worth its cost: 28 of 30 best PSNR/SSIM entries with fewer than one million
parameters, with the advantage concentrated where the account predicts. We have
also reported where it does not hold, and deployed the model on a
Raspberry~Pi~5 to confirm it runs within the envelope of a credit-card
computer.

Two directions follow. Perceptually oriented training via adversarial objectives
\cite{goodfellow2014gan,ledig2017srgan,wang2018esrgan} could address the
smoothness of complex texture, evaluated with perceptual metrics
\cite{zhang2018lpips,ke2021musiq,wang2023clipiqa} alongside fidelity ones given
the perception--distortion trade-off \cite{blau2018perception}. And since we
train on bicubic-downsampled pairs, extending to real-world degradation
\cite{wang2021realesrgan} would test whether the account generalises beyond
synthetic data.

\section*{CRediT authorship contribution statement}

\textbf{Chih-Hsiang Yang}: Conceptualization, Methodology, Software,
Validation, Formal analysis, Investigation, Data curation, Visualization,
Writing -- original draft. \textbf{Chia-Min Lin}: Methodology, Validation,
Investigation, Writing -- review \& editing. \textbf{Ching-Yu Tsai}:
Validation, Investigation, Writing -- review \& editing.
\textbf{Yung-Che Wang}: Validation, Investigation, Writing -- review \&
editing. \textbf{Jen-Shiun Chiang}:
Conceptualization, Methodology, Resources, Supervision, Project
administration, Funding acquisition, Writing -- review \& editing.

\section*{Declaration of competing interest}

The authors declare that they have no known competing financial interests or
personal relationships that could have appeared to influence the work reported
in this paper.

\section*{Data availability}

All datasets used in this study (DIV2K, Flickr2K, Set5, Set14, BSD100,
Urban100 and Manga109) are publicly available. Code and trained models will be
made available upon reasonable request.

\section*{Acknowledgements}

The authors thank the members of the laboratory for their assistance with
experimental resources and for helpful discussions.

\bibliographystyle{elsarticle-num}
\bibliography{sfmformer}

\end{document}